\documentclass[10pt,twocolumn,letterpaper]{article}

\usepackage{cvpr}              

\usepackage{graphicx}
\usepackage{amsmath}     
\usepackage{multirow}
\usepackage{amssymb}
\usepackage{colortbl}
\usepackage{wrapfig}
\usepackage[most]{tcolorbox}
\usepackage{makecell}
\usepackage{xspace}
\usepackage{subcaption}
\usepackage{enumitem}
\usepackage{pifont}
\usepackage{color, colortbl}
\usepackage{algorithm}
\usepackage{algorithmicx}
\usepackage{algpseudocode}

\newcommand{\bl}{Visual-RFT\xspace}
\newcommand{\ours}{ReFine-RFT\xspace}

\newcommand{\ao}{\textit{Answer-only}\xspace}
\newcommand{\cott}{\textit{CoT}\xspace}
\newcommand{\alg}{MRN\xspace}
\newcommand{\think}{\texttt{<think>...</think>}\xspace}
\newcommand{\answer}{\texttt{<answer>...</answer>}\xspace}

\newcounter{takeawayonly}
\newcommand{\parag}[1]{\vspace{+0.0mm}\noindent\textbf{#1}}
\newcommand{\takeawayonly}[1]{
    \vspace{-0.05cm}
    \refstepcounter{takeawayonly}
    \begin{tcolorbox}[
        colback=darkgreen!8,                       
        colframe=darkgreen!95,                     
        arc=4pt,                    
        boxsep=5pt,                 
        left=2pt,                  
        right=2pt,                 
        top=4pt,                    
        bottom=4pt,                 
        boxrule=0.8pt,              
        drop shadow=gray!30!white,  
        enhanced jigsaw             
    ]
    \vspace{-0.15cm}
        \parag{\textbf{\textit{Finding\,\thetakeawayonly:}}} #1
    \vspace{-0.15cm}
    \end{tcolorbox}
}
\newcounter{conclusion}
\newcommand{\conclusion}[1]{
    \vspace{-0.05cm}
    \refstepcounter{conclusion}
    \begin{tcolorbox}[
        colback=darkgreen!8,                       
        colframe=darkgreen!95,                     
        arc=4pt,                    
        boxsep=5pt,                 
        left=2pt,                  
        right=2pt,                 
        top=4pt,                    
        bottom=4pt,                 
        boxrule=0.8pt,              
        drop shadow=gray!30!white,  
        enhanced jigsaw             
    ]
    \vspace{-0.15cm}
        \parag{\textbf{\textit{Conclusion\,\theconclusion:}}} #1
    \vspace{-0.15cm}
    \end{tcolorbox}
}

\renewcommand{\paragraph}[1]{\vspace{1.25mm}\noindent\textbf{#1}}

\def\eg{\emph{e.g}\onedot}

 \def\vs{\emph{vs}\onedot}

\makeatother

\definecolor{green}{HTML}{0aa344}
\definecolor{red}{HTML}{c93756}
\definecolor{darkgreen}{HTML}{068C52}
\definecolor{fullgreen}{rgb}{0.502, 0.788, 0.643}
\definecolor{fullred}{rgb}{0.800, 0.447, 0.541}
\definecolor{lightgreen}{RGB}{225, 239, 217}   
\definecolor{lightblue}{RGB}{203, 220, 235}   
\definecolor{fullgray}{RGB}{219, 223, 234}   
\definecolor{fullpurple}{RGB}{205, 193, 255}
\definecolor{darkred}{RGB}{204, 114, 138}
\definecolor{darkpurple}{RGB}{171, 151, 255}
\definecolor{darkgray}{RGB}{114, 114, 114}
\definecolor{teal}{HTML}{14B8A6}
\definecolor{sky}{HTML}{38BDF8}
\definecolor{indigo}{HTML}{6366F1}
\definecolor{navy}{HTML}{1E3A8A}
\definecolor{amber}{HTML}{F59E0B}
\definecolor{coral}{HTML}{FF6B6B}
\definecolor{peach}{HTML}{FFB4A2}
\definecolor{sage}{HTML}{A8C3A1}
\definecolor{dustyblue}{HTML}{9BB4C7}
\definecolor{mauve}{HTML}{BFA6C9}
\definecolor{clay}{HTML}{C9B29B}

\usepackage{microtype}

\definecolor{cvprblue}{rgb}{0.21,0.49,0.74}
\usepackage[pagebackref,breaklinks,colorlinks,allcolors=cvprblue]{hyperref}

\def\paperID{} 
\def\confName{CVPR}
\def\confYear{2026}

\title{Can Textual Reasoning Improve the Performance of MLLMs on Fine-grained Visual Classification?}

\author{
Jie Zhu\textsuperscript{1}\quad Yiyang Su\textsuperscript{1}\quad Xiaoming Liu\textsuperscript{1,2} \\
\textsuperscript{1}Michigan State University \quad \textsuperscript{2}University of North Carolina at Chapel Hill\\
{\tt\small \{zhujie4, suyiyan1\}@msu.edu \quad liuxm@cs.unc.edu }
}

\begin{document}
\maketitle

\begin{abstract}
Multi-modal large language models (MLLMs) exhibit strong general-purpose capabilities, yet still struggle on Fine-Grained Visual Classification (FGVC), a core perception task that requires subtle visual discrimination and is crucial for many real-world applications. A widely adopted strategy for boosting performance on challenging tasks such as math and coding is Chain-of-Thought (CoT) reasoning. However, several prior works have reported that CoT can actually harm performance on visual perception tasks. These studies, though, examine the issue from relatively narrow angles and leave open why CoT degrades perception-heavy performance. We systematically re-examine the role of CoT in FGVC through the lenses of zero-shot evaluation and multiple training paradigms. Across these settings, we uncover a central paradox: the degradation induced by CoT is largely driven by the reasoning length, in which longer textual reasoning consistently lowers classification accuracy. We term this phenomenon the ``Cost of Thinking''. Building on this finding, we make two key contributions: (1)  \alg, a simple and general plug-and-play normalization method for multi-reward optimization that balances heterogeneous reward signals, and (2) ReFine-RFT, a framework that combines ensemble rewards with \alg to constrain reasoning length while providing dense accuracy-oriented feedback. Extensive experiments demonstrate the effectiveness of our findings and the proposed ReFine-RFT, achieving state-of-the-art performance across FGVC benchmarks. Project page: \href{https://refine-rft.github.io/}{ReFine-RFT}.

\end{abstract}
\begin{figure}[t!]
    \centering
    \includegraphics[width=\linewidth]{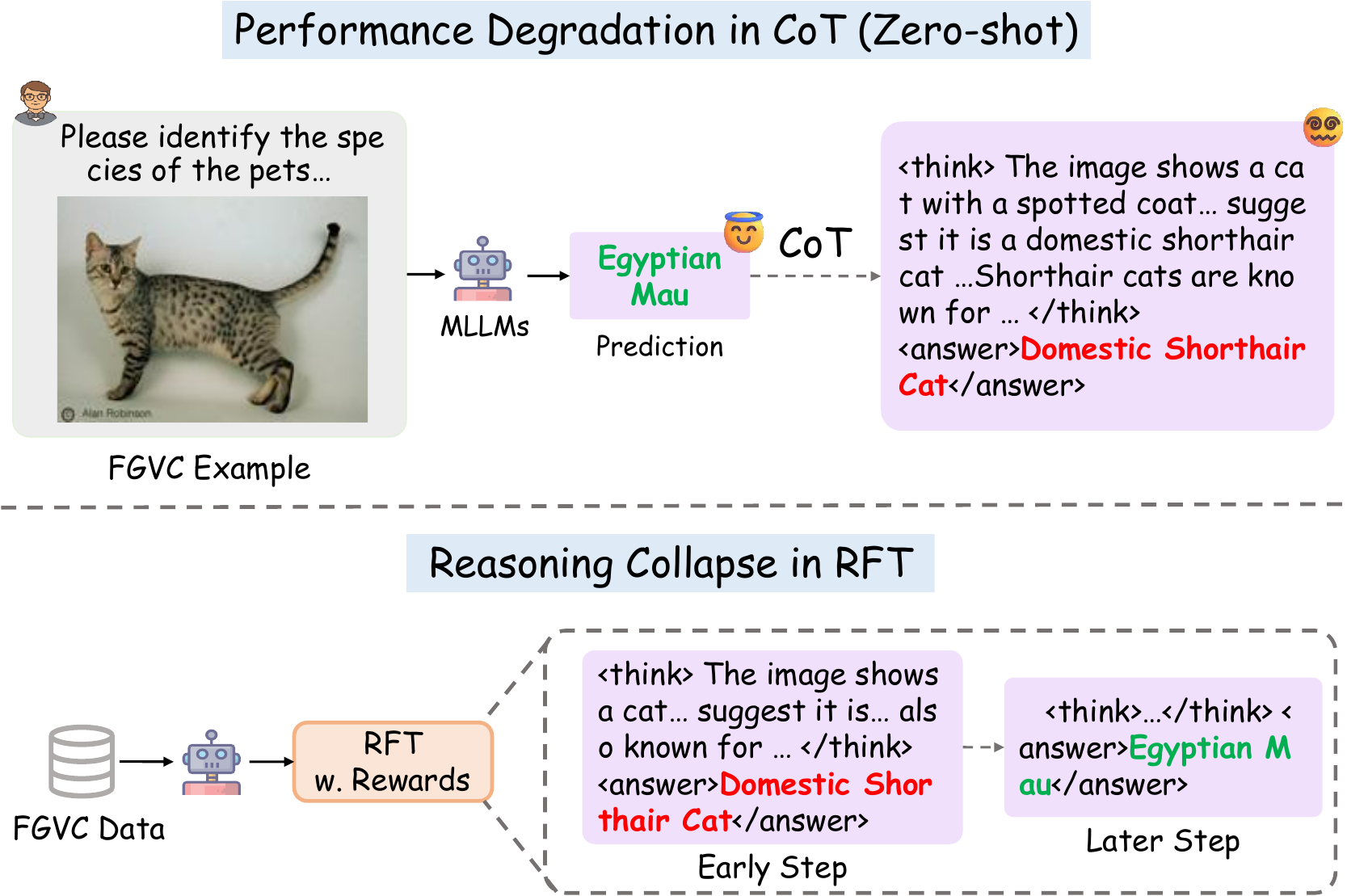}
    \caption{\textbf{Performance degradation with CoT and reasoning collapse in RFT.} In zero-shot evaluation (top), MLLMs predict the correct label directly, but adding CoT reasoning leads to a wrong answer. During RFT (bottom), reasoning length steadily shrinks while accuracy improves, indicating a reasoning collapse.}
    \label{fig:motivation}
\vspace{-1em}
\end{figure}

\section{Introduction} \label{sec:intro}

Multi-modal large language models (MLLMs) have demonstrated remarkable capabilities in general vision-language understanding, enabling seamless interaction across images and text and driving progress toward versatile, general-purpose AI systems~\cite{bai2023qwen, bai2025qwen2, zhu2025internvl3}. As these models are increasingly deployed as unified interfaces for perception and reasoning, their ability to handle fine-grained visual understanding becomes particularly critical~\cite{he2025analyzing}. Fine-grained Visual Classification (FGVC) requires discriminating among subordinate-level categories that exhibit only subtle visual differences (\eg, car models or plant varieties) and serves as a foundation for more advanced perception-centric tasks such as object-centric visual question answering~\cite{zhang2024visually, zhang2023tile, geigle2024african, he2025analyzing}. For example, a model unable to reliably differentiate between similar-looking pet breeds (\eg, golden retriever \vs labrador) is also likely to fail in answering follow-up questions about their behavioral traits or health conditions. Unlike other recognition tasks~\cite{fu2025learning, zhang2026unleashing, huang2026unlocking, zhang2026towards, su2025hamobe}, FGVC demands precise visual perception and sensitivity to subtle cues such as fur texture, fine-grained shape differences, or minor pattern variations.
Studying FGVC in the context of MLLMs directly probes their visual grounding and fine-grained feature extraction abilities. This lets us assess whether MLLMs can act as trustworthy assistants in visually demanding domains (\eg, ecology, medical imaging, industrial inspection)~\cite{zhang2024visually}.

Despite their sophisticated architectures, current MLLMs exhibit clear limitations on FGVC, often failing to capture the subtle distinctions that define fine-grained categories~\cite{he2025analyzing, zhang2024visually, geigle2024african}, especially in few-shot and open-ended scenarios where training data is scarce and models must adapt to specialized domains from limited supervision without overfitting. This naturally raises the question of whether the textual description can help compensate for these perceptual weaknesses. A widely held belief in the community is that eliciting Chain-of-Thought (CoT) reasoning improves performance on complex tasks such as math and coding~\cite{wei2022chain, zhang2022automatic,zhang2024chain, shao2024deepseekmath, hui2024qwen2, sui2025stop}, and recent visual-oriented frameworks such as Visual-RFT~\cite{liu2025visual} introduce CoT to steer MLLMs toward enhanced visual perception capabilities, achieving state-of-the-art performance on FGVC. However, several prior works have shown that explicit textual reasoning can paradoxically \emph{reduce} accuracy compared to direct predictions~\citep{liu2024mind, tam2024let, jiang2025mme, li2025think}. These studies, however, though informative, generally examine only limited settings: either focusing solely on zero-shot evaluation or comparing CoT and answer-only predictions at a coarse level. This motivates us to systematically re-examine the role of textual reasoning from broader and in-depth evaluation and training perspectives. Specifically, we formulate a key research question:

\begin{tcolorbox}[colback=lightgray!55, colframe=lightgray!75!black, boxrule=0pt, arc=2mm, left=1mm, right=1mm, top=1mm, bottom=1mm]
\textit{Is textual reasoning detrimental to fine-grained visual perception, or do current methods simply employ it in a suboptimal way?}
\end{tcolorbox}

\noindent To answer this question, we conduct a comprehensive investigation through two aspects: i) zero-shot scenario, and ii) using a Reinforcement Fine-Tuning (RFT) framework to diagnose and manage this trade-off between reasoning and visual accuracy. Our diagnostic experiments reveal several key observations: First, CoT harms zero-shot FGVC performance, shown in Fig.~\ref{fig:motivation} (top) and Tab.~\ref{tab:performance_drop_cot}; second, Reasoning Collapse in RFT, shown in Fig.~\ref{fig:motivation} (bottom) and Fig.~\ref{fig:rft_length_curve}, where MLLMs gradually learn to suppress verbose reasoning while optimizing for accuracy during RFT. Our in-depth analysis not only corroborates previous observations but further extends them by revealing a central insight: \emph{the length of textual reasoning itself is a critical factor for fine-grained visual perception.} We observe a consistent negative correlation between reasoning length and accuracy: the longer the reasoning content, the worse the visual perception performance. We term these the \textbf{``Cost of Thinking''}, which reveals that fine-grained visual perception tasks might benefit from concise rather than elaborate reasoning for MLLMs.

Based on our findings, we introduce \textbf{\ours}, a novel RFT framework designed to constrain reasoning and improve accuracy. Our framework features two key technical innovations to solve the core challenges of this task. First, to overcome the sparse and semantically naive signal of binary accuracy rewards, we introduce an \textbf{ensemble, semantically-aware reward}, which provides a dense and continuous learning signal while explicitly restricting the reasoning length. Second, to optimize multi-objective reward space, we propose \textbf{Multi-reward Normalization (\alg)}, a plug-and-play module that stabilizes training by reducing and smoothing the variance of the reward signals. Our contributions are fourfold:

\begin{itemize}[noitemsep, topsep=2pt, leftmargin=*]
    \item We empirically characterize the \textbf{``Cost of Thinking''} in FGVC, showing that verbose CoT systematically degrades MLLM performance on fine-grained perception tasks.
    \item We propose \textbf{\alg}, a plug-and-play normalization that independently normalizes heterogeneous reward signals in multi-objective settings.
    \item We introduce \textbf{\ours}, an RFT framework that integrates \alg with ensemble rewards to primarily optimize accuracy while controlling reasoning length.
    \item \ours achieves state-of-the-art across multiple FGVC benchmarks, validating the effectiveness of our findings and the proposed method.
\end{itemize}
\section{Related Works} \label{sec:related_works}

\paragraph{Reasoning Ability of MLLMs.}
Prior works have demonstrated that reasoning could improve performance on complex tasks like math and coding~\cite{wei2022chain, nye2021show, zhang2023automatic, zhang2023multimodal, li2026toward, shen2025fine, liu2026palm}. However, empirical evidence~\cite{liu2024mind, yu2025perception, jiang2025mme, sprague2024cot} shows that CoT reasoning can introduce spurious explanations and degrade visual perception accuracy. For example, No-Thinking-RFT~\cite{li2025think} presents that visual tasks do not need thinking. However, it primarily compares performance between training with CoT and answer-only prompts. We systematically re-evaluate textual reasoning for fine-grained perception across zero-shot and different training regimes, revealing that thinking length is the key to ``Cost of Thinking’’. 

\paragraph{Fine-grained Visual Classification in MLLMs.} 
FGVC~\cite{wah_branson_welinder_perona_belongie_2011, wei2021fine, maji2013fine, krause20133d, nilsback2008automated, zhu2025quality, liu2025person, parkhi2012cats, xu2026emotag} focuses on subcategory-level recognition that requires capturing subtle visual cues. With the advent of MLLMs, recent works~\cite{ren2023chatgpt, tong2024eyes, cui2024fine, liu2024democratizing, chen2023atm, he2025analyzing, guo2026holistic, deepfakeagent2026guo} explore prompting and adaptation strategies to improve the FGVC of MLLMs. Visual-RFT~\cite{liu2025visual} further applies RFT with CoT reasoning and achieves additional gains. Building on our findings, we propose \ours, which combines ensemble rewards with \alg to explicitly constrain textual reasoning while further enhancing the fine-grained visual perception capability of MLLMs.

\begin{table*}[t!]
\tabcolsep=0.1cm
\centering
\resizebox{0.9\linewidth}{!}{%
\begin{tabular}{lcccccccccc}
    \toprule
    \multirow{3}{*}{\bf Model}  & \multicolumn{2}{c}{\textbf{Aircrafts-102}} & \multicolumn{2}{c}{\textbf{Flowers-102}}& \multicolumn{2}{c}{\textbf{Cars-196}} & \multicolumn{2}{c}{\textbf{Pets-37}} & \multicolumn{2}{c}{\textbf{Average}}  \\
    \cmidrule(lr){2-3} \cmidrule(lr){4-5} \cmidrule(lr){6-7} \cmidrule(lr){8-9} \cmidrule(lr){10-11}
    & \ao & \cott & \ao  & \cott & \ao  & \cott & \ao  & \cott & \ao & \cott \\
    \midrule
    \rowcolor{gray!30} \multicolumn{11}{c}{\textit{Open-source Non-reasoning Models}} \\
    Qwen2-VL-2B & 47.5 & 45.9 & 55.7 & 54.8 & 82.6 & 56.8 & 56.4 & 66.4 & 60.5 &  55.9  \\
    Qwen2-VL-7B & 53.5 & 42.3 & 55.8 & 51.1 & 83.9 & 76.5 & 51.4 & 61.1 & 61.2 & 57.8  \\
    Qwen2.5-VL-7B & 54.0 & 41.7 & 51.1 & 36.3 & 73.0 & 66.9 &  52.5 & 62.4& 57.7 & 51.8  \\
    InternVL2.5-8B & 13.8 & 11.9 & 20.1 & 12.9 & 33.5 & 31.9 & 48.1 & 50.4 & 28.9 & 26.8  \\
    InternVL3-8B & 14.2 & 14.5 & 23.2 & 10.1 & 42.2 & 36.5 & 37.6 & 42.8 & 29.3 & 25.9   \\
    \rowcolor{gray!30} \multicolumn{11}{c}{\textit{Open-source Reasoning Models}} \\
    R1-OneVision-7B-RL & - & 42.0 & - & 59.2 & - & 49.3 & - & 68.8 & - &  54.8 \\
    \bottomrule
    \end{tabular}
    }
\caption{\textbf{Performance Degradation of CoT on FGVC benchmarks.} 
We evaluate several open-source MLLMs on four FGVC datasets under two prompting settings: \ao and \textit{Chain-of-Thought (CoT)}. Results show notable performance degradation when CoT reasoning is applied. For R1-OneVision-7B-RL, \ao are omitted, as it generates CoT-style outputs even under \ao.}
\label{tab:performance_drop_cot}
\vspace{-1em}
\end{table*}  

\paragraph{Reinforcement Fine-tuning.} Reinforcement Learning (RL) originated in control theory for optimal decision-making in dynamic environments~\cite{sutton1998reinforcement, kaelbling1996reinforcement, zhu2024fairness, matsuo2022deep, han2023survey, xu2025stare}. Recent research demonstrates that RL can significantly enhance the reasoning and problem-solving capabilities of LLMs and MLLMs compared with Supervised Fine-tuning (SFT)~\cite{guo2025deepseek, bai2022training, sun2023aligning, shao2024deepseekmath, hui2024qwen2}. DeepSeek-R1~\cite{guo2025deepseek} introduces Group Relative Policy Optimization (GRPO), substantially improving reasoning and generalization. Follow-up works further apply GRPO to other tasks such as visual grounding~\cite{shen2025vlm, huang2025vision, liu2025visual, tan2025reason, zhang2025r1, chen2025suitability, peng2025lmm, yang2025r1, zhu2026fusionagent}, typically using simple rule-based signals such as accuracy. However, existing methods ignore heterogeneity across reward functions. We propose \alg to balance multi-reward signals, and use an ensemble of rewards to provide denser reward feedback.
\section{Cost of Thinking in FGVC} \label{sec:preliminary}

\subsection{Experiment Setup} \label{subsec:experiment_setup}

\paragraph{Datasets.}
We select widely adopted FGVC benchmarks: FGVC-Aircraft~\citep{maji2013fine}, Stanford-Cars~\citep{krause20133d}, Flowers-102~\citep{nilsback2008automated}, and Oxford-Pets~\citep{parkhi2012cats}. Considering a real-world scenario where fine-grained labeled data might be scarce, we use a 4-shot dataset provided by~\cite{liu2025visual}. We perform FGVC as an open-ended QA task to mimic the real-world application.

\paragraph{Models and Prompts.}
We evaluate \ao and \cott prompts across several open-source MLLMs: Qwen2/2.5-VL series~\citep{wang2024qwen2, bai2025qwen2}, InternVL series~\citep{chen2024expanding, zhu2025internvl3}, and the reasoning model R1-OneVision~\citep{yang2025r1}. For RFT training, we use Qwen2-VL-2B~\cite{wang2024qwen2} as the base model. We use the \cott prompt from~\cite{liu2025visual}, and the following \ao prompt as an example for Flowers-102:

\begin{tcolorbox}[colback=lightblue!45, colframe=lightblue!85!black, boxrule=0pt, arc=2mm, left=1mm, right=1mm, top=1mm, bottom=1mm]
\textit{\quad This is an image containing a plant. Please identify the species of the plant based on the image. Only provide the \textbf{final answer directly}, without any explanation or special formatting.}
\end{tcolorbox}

\paragraph{Reward Functions for RFT.}
We follow \bl~\cite{liu2025visual} and use format reward $R_f(o_i)$ and accuracy reward $R_{cls}(a_i, y)$ to improve the instruction-following capability and answer accuracy. The format reward is a binary signal that enforces strict adherence to the required structured output template, assigning 1 if the model’s response $o_i$ correctly follows the sequential \think and \answer tags, and 0 otherwise. The classification reward $R_{cls}(a_i, y)$ measures prediction accuracy based on the class encoded within the \answer tags, yielding 1 when the the ground-truth label $y$ is in the predicted label $a_i$ extracted from the answer tags of $o_i$ and 0 otherwise; 

To investigate the effects of thinking length, we introduce a thinking length reward $R_{len}(o_i)$ that assigns a binary score based on whether the thinking length lies within a specified range. 
Let $t_i$ denote the reasoning content extracted from the model output $o_i$, and let $L_i = \lvert t_i \rvert$ be its character length. 
Given predefined bounds $(L_{\min}, L_{\max})$, the reward is computed as:
\begin{equation}
\label{eq:thinking-length-reward}
R_{len}(o_i) =
\begin{cases}
1, & \text{if } L_{\min} \le L_i \le L_{\max}, \\[6pt]
0, & \text{otherwise.}
\end{cases}
\end{equation}
This formulation encourages the model to produce reasoning traces whose lengths fall within the desired interval, enabling explicit control over the degree of internal deliberation. When the \texttt{<think>} tags are missing or improperly formatted, the reward is set to 0.

\begin{figure*}[t!]
    \centering
    \includegraphics[width=0.9\linewidth]{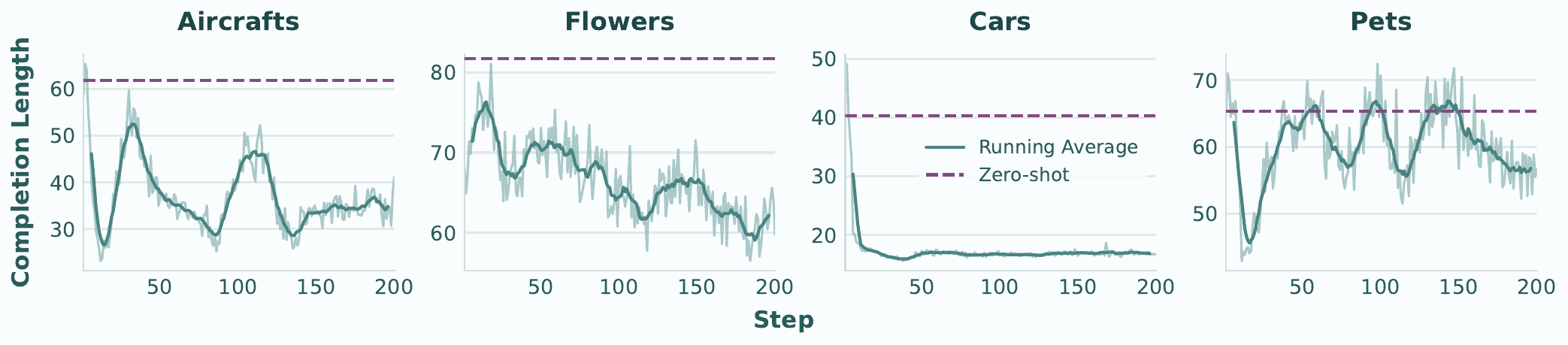}
    \caption{\textbf{Dynamics of reasoning length during RFT across FGVC datasets.} The dark green lines denote the running average of completion lengths throughout RFT FGVC tasks. Across all datasets, the reasoning content length rapidly decreases and stabilizes at a shorter range, suggesting that RFT discourages excessive reasoning generation and promotes concise, decision-focused responses. [Zero-shot: average content length of base model on the evaluation set; Step: the cumulative number of gradient update steps.]}
    \label{fig:rft_length_curve}
\end{figure*}

\subsection{Results \& Findings}

\paragraph{Performance Degradation in CoT under Zero-Shot.}
As shown in Tab.~\ref{tab:performance_drop_cot}, incorporating Chain-of-Thought (CoT) prompting consistently leads to accuracy degradation across all FGVC datasets. Non-reasoning models exhibit an average drop of 3–6\% when switching from \ao to \cott prompts, while even reasoning-oriented model, R1-OneVision, achieves only moderate accuracy under \cott and still generates \cott response for \ao. This indicates that visual reasoning chains often introduce useless or hallucinatory explanations rather than improving decision quality. ~\cite{liu2024mind, tam2024let, jiang2025mme} also reveal similar phenomena that reasoning might be harmful to visual recognition.
However, this observation raises a fundamental question: \textit{Is reasoning intrinsically harmful to visual perception tasks, or is the degradation simply a byproduct of zero-shot misalignment between CoT prompting and model training?} We further explore how reasoning evolves under RFT, as models adapt their generation strategy through reward-driven learning. 

\paragraph{Reasoning Collapse in RFT.}
To probe the dynamics of reasoning adaptation, we track the change of completion length throughout RFT. We set up the experiment following Visual-RFT~\cite{liu2025visual} with format reward and classification reward described in Sec.~\ref{subsec:experiment_setup}. As shown in Fig.~\ref{fig:rft_length_curve}, the average reasoning length exhibits a consistent downward trend across all FGVC datasets. At the early stages of RFT, model outputs are verbose and exploratory, but as training progresses, the content length rapidly declines and stabilizes at a compact range. Notably, the final completion lengths after RFT are shorter than those in the zero-shot setting. We refer to this phenomenon as \textit{reasoning collapse}: an emergent behavior where RFT implicitly discourages long reasoning chains for visual perception tasks, optimizing instead for concise, confident answer prediction. This collapse suggests that the model learns to suppress reasoning steps that do not contribute to reward maximization, thereby aligning its behavior more closely with discriminative objectives. In other words, RFT appears to regularize the reasoning process itself, favoring precision and efficiency over verbosity and exploration, a tendency that aligns with findings from~\cite{li2025think}.

However, this behavior may also result from reward hacking, since no explicit constraint is imposed on the reasoning process, leading the MLLM to generate only minimal reasoning content. Building upon this observation, we design the subsequent experiments to further verify and quantify the effect for reasoning.

\begin{figure}[t!]
    \centering
    \includegraphics[width=\linewidth]{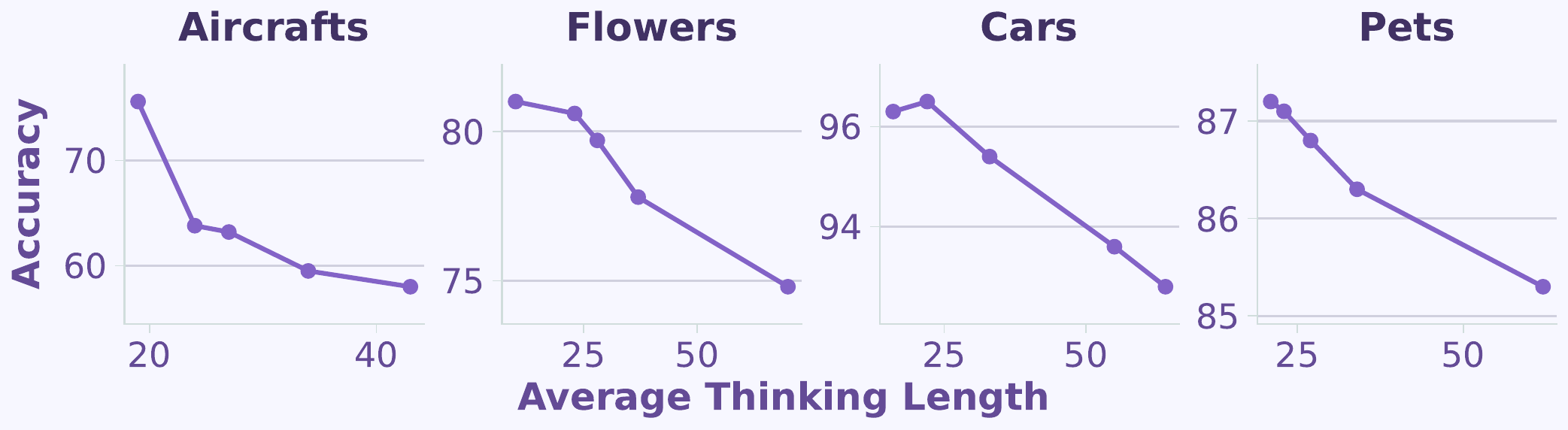}
    \caption{\textbf{Impact of reasoning length on FGVC performance.} We analyze the relationship between average reasoning (thinking) length and classification accuracy across FGVC datasets. As the average thinking length increases, performance consistently declines, indicating that excessive reasoning generation introduces noise or distracting the model from key discriminative visual cues.}
    \label{fig:thinking_length_effect}
\vspace{-1em}
\end{figure}

\begin{figure*}[t!]
    \centering
    \includegraphics[width=0.9\linewidth]{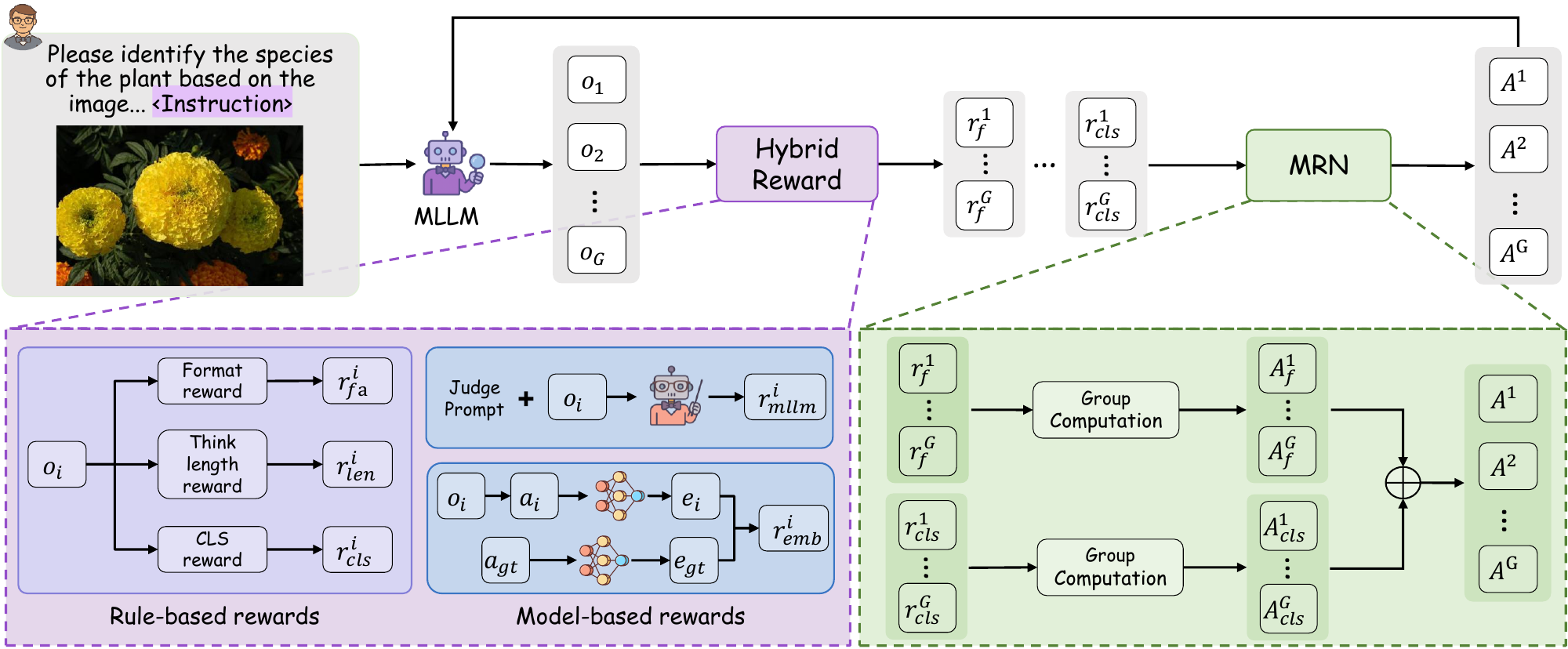}
    \caption{\textbf{Overview of \ours.} Given a question, the model generates multiple candidate responses, each evaluated using an ensemble reward that combines rule-based rewards and model-based rewards like MLLM-based accuracy reward and embedding similarity reward. The proposed \alg then normalizes the rewards for each function to compute the final advantages used to update the MLLM.}
    \label{fig:methods}
\vspace{-.5em}
\end{figure*}

\paragraph{Effects of Thinking Length in RFT.}
To further examine whether the reasoning length is beneficial or detrimental, we explicitly manipulate the reasoning process by involving the thinking length reward during RFT. We gradually limit the reasoning content length from $[0,20]$ to $[60,80]$. As shown in Fig.~\ref{fig:thinking_length_effect}, enforcing longer reasoning sequences leads to a clear decline in classification accuracy across all FGVC datasets. This inverse correlation demonstrates that extended reasoning is not only unhelpful but can actively degrade performance by introducing textural reasoning into the responses. In contrast, shorter reasoning traces yield higher accuracy, indicating that optimal visual performance is achieved with minimal reasoning and precise visual perception and localization. However, the degradation may also stem from low-quality CoT, as reasoning quality is unsupervised during RFT. This motivates analyzing whether higher-quality CoT can close this gap.

\paragraph{Answer-only Surpasses CoT in SFT.} To analyze the effects of CoT quality, we use GPT-4o~\cite{hurst2024gpt} to generate the high-quality CoT data for SFT-CoT training. As shown in Tab.~\ref{tab:main_results}, SFT-AO consistently outperforms SFT-CoT, indicating that the degradation is not simply due to the quality of CoT. This finding complements our “Cost of Thinking’’ analysis from the SFT perspective, showing that excessive reasoning can harm fine-grained visual perception in both training and inference. Taken together with the observations under zero-shot and RFT, these results reveal a key finding:

\takeawayonly{For fine-grained visual tasks, \textbf{thinking length is the key factor}: excessive reasoning hurts performance, and MLLMs benefit more from \textbf{concise} responses than from elaborate reasoning.} \label{finding1}
\vspace{-.7em}

\section{Methods} \label{sec:methods}

Inspired by our findings, we propose \ours, which improves RFT by combining the ensemble reward with a Multi-Reward Normalization scheme. The ensemble reward jointly constrains reasoning length and provide dense accuracy feedback, while Multi-Reward Normalization stabilizes optimization across heterogeneous reward signals. An overview of \ours is shown in Fig.~\ref{fig:methods}.

\subsection{Ensemble Reward}

The ensemble reward is composed of the format, accuracy, and thinking-length rewards defined in Sec.~\ref{subsec:experiment_setup}, together with two complementary rewards: an MLLM-based accuracy reward and an embedding similarity reward, which jointly provide a richer and accuracy-centric feedback.

\paragraph{MLLM-based Accuracy Reward.}
The classification reward provides only binary supervision through exact string matching between the predicted and ground-truth answers, which fails to capture semantic similarity. 
For example, predictions such as ``\textit{Datura stramonium}'' vs.~``\textit{thorn apple}'' denote the same subcategory but would be penalized under hard matching, and ``\textit{Dodge Dakota}'' vs.~``\textit{2007 Dodge Dakota Club Cab}'', which is missing some details. To provide a smoother and more semantically-aware signal, we introduce an MLLM-based Accuracy Reward $R_{mllm}$ that employs an MLLM as a teacher to grade each prediction. Given a predicted answer and the reference label, the MLLM is prompted to output a score from 0 to 10 based on its semantic alignment, which is then normalized to $[0,1]$. This continuous reward function provides fine-grained feedback by assigning high scores to fully correct answers, intermediate scores to semantically similar ones, and low scores to irrelevant predictions. To mitigate potential scoring biases of the reward model~\cite{shen2023loose, lambert2025rewardbench}, we include few-shot grading examples in the prompt and use this reward together with the embedding similarity reward.

\paragraph{Embedding Similarity Reward.}
To further provide a smooth and continuous supervision signal, we introduce an embedding similarity reward $R_{emb}$ that measures the semantic closeness between the predicted and ground-truth answers in an embedding space. 
Given the extracted predicted answer $a_i$ from the \texttt{<answer>} tags of the model output and the reference label, both are encoded into text embeddings using a pretrained text embedding model. 
The cosine similarity between the predicted embedding $\mathbf{e}_{i}$ and the ground-truth embedding $\mathbf{e}_{gt}$ is used as the reward:
\[
R_{emb} = \cos(\mathbf{e}_{i}, \mathbf{e}_{gt}) \in [0,1].
\]
This continuous reward provides a differentiable measure of semantic alignment, encouraging the model to produce answers that are semantically close to the reference even when lexical forms differ.

\subsection{Multi-reward Normalization (\alg)} 
As shown in Fig.~\ref{fig:methods}, for a given question $q$ and image $x$, GRPO requires the model to sample $G$ diverse responses ${o_1, o_2, \dots, o_G}$ from the current model $\pi_{\theta}$ and obtains final rewards ${r^1, r^2, \dots, r^G}$ for ${o_1, o_2, \dots, o_G}$, respectively. GRPO assesses the relative quality by normalizing $r^i$ using the mean and standard deviation of the group reward:
\begin{equation}
A^i=\frac{r^i-\mathrm{mean}({r^1,\ldots,r^G})}{\mathrm{std}({r^1,\ldots,r^G})},
\end{equation}
where $A^i$ denotes the advantage of the $i$-th response. With the group normalization, GRPO encourages the model to sample preferred answers with a higher reward.

\begin{figure}[t!]
    \centering
    \includegraphics[width=\linewidth]{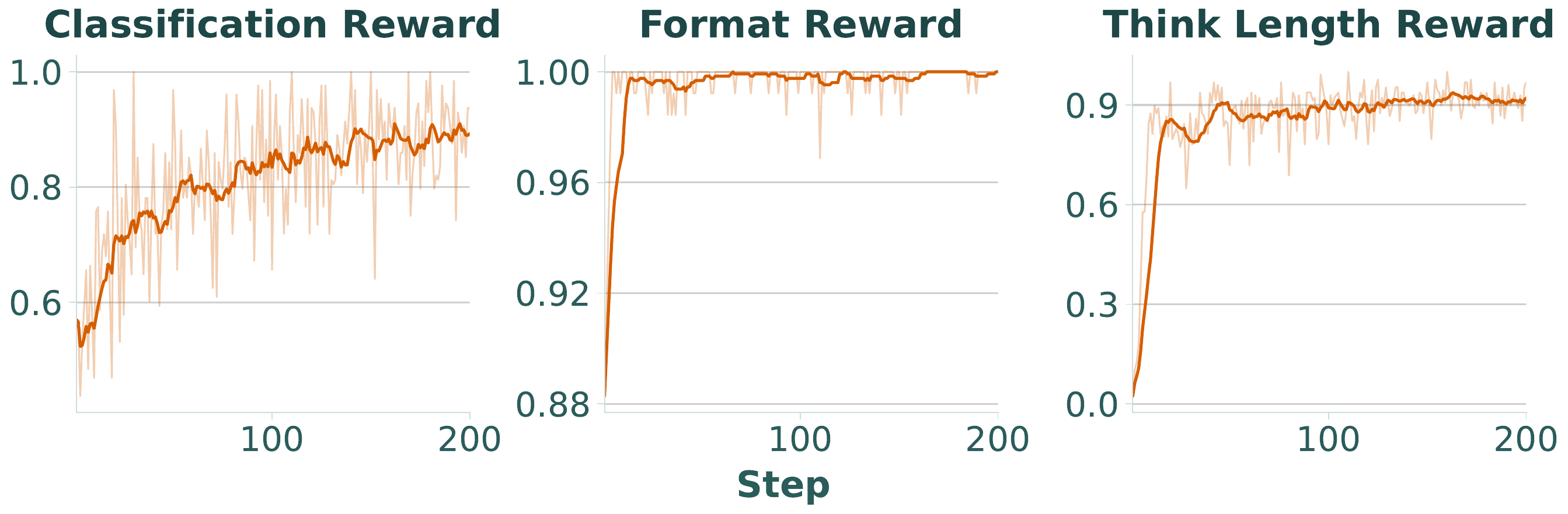}
    \caption{\textbf{Differences among rewards during training.} Each reward exhibits distinct convergence speed, value range, and saturation point, reflecting the heterogeneity of different rewards.}
    \label{fig:rewards_difference_comparison_pets}
\vspace{-.5em}
\end{figure}

In practical training scenarios, multiple reward signals (\eg, format and classification) are often combined to guide optimization. In the original GRPO, these heterogeneous rewards are first aggregated into a single scalar value before performing group normalization. However, in reality, different rewards exhibit varying levels of difficulty and convergence rates. As shown in Fig.~\ref{fig:rewards_difference_comparison_pets}, the format reward may quickly saturate in early training, thereby dominating the total reward and diluting the influence of other, more informative rewards such as accuracy. To address this issue, we propose Multi-reward Normalization (\alg), which performs group normalization independently for each reward component before aggregation. Specifically, given $K$ reward types ${r_{(1)}^i, r_{(2)}^i, \ldots, r_{(K)}^i}$ for the $i$-th response $o_i$, we compute the normalized advantage for each reward as:
\begin{equation}
A_{(k)}^i = \frac{r_{(k)}^i - \mathrm{mean}({r_{(k)}^1, \ldots, r_{(k)}^G})}{\mathrm{std}({r_{(k)}^1, \ldots, r_{(k)}^G})},
\end{equation}
and then aggregate them to obtain the final advantage:
\begin{equation}
A^i = \sum_{k=1}^{K} A_{(k)}^i.
\end{equation}
This normalization places all rewards on a comparable scale, leading to a more stable and balanced optimization. The pseudocode is provided in Alg.~\ref{alg:advantage_norm}.

\begin{table*}[t!]
\tabcolsep=0.1cm
\centering
\resizebox{1\linewidth}{!}{
\begin{tabular}{l|c|c|cccc|c}
\toprule
{\textbf{Methods}} & {\textbf{FT Methods}} & {\textbf{FT Types}} & {\textbf{Aircrafts-102}} &{\textbf{Flowers-102}} &{\textbf{Cars-196}} & {\textbf{Pets-37}} &{\textbf{Average}} \\
\midrule
Qwen2-VL-2B~\cite{wang2024qwen2}  & \textit{Zero-shot} & \textit{-} & 45.9 & 54.8 & 56.8 & 66.4 & 56.0 \\
Finedefics-8B~\cite{he2025analyzing} & \textit{SFT} & \textit{Fully-FT} & 63.8 &  89.9 &  84.7 & 92.2 & 82.7 \\
\midrule
SFT-AO            & \multirow{3}{*}{\textit{SFT}} & \textit{Fully-FT} & 67.9 & 58.5 &  40.5 & 55.5 & 55.6 \\
SFT-AO            & &  \textit{Lora} & 78.3 & 74.8 & 80.0 & \underline{87.6} & 80.2 \\
SFT-CoT           & &  \textit{Lora} & 73.9 & 74.4 &  52.3  & 87.5 & 72.0 \\
\midrule
Visual-RFT~\cite{liu2025visual}         & \multirow{5}{*}{\textit{RFT}}  & \textit{Fully-FT} &  74.8 & 71.4 & 95.3 & 86.1 & 81.9 \\
Visual-RFT~\cite{liu2025visual}         & &  \textit{Lora} &  75.6 & 74.1 & \underline{95.7} & 86.0 & 82.9 \\
No-Thinking-RFT~\cite{li2025think}      & & \textit{Fully-FT} & - & 71.2 & - & 86.1 & -  \\

\rowcolor{fullgreen!30} \bf \ours-AO (Ours) &  
 &  \textit{Lora} & \underline{78.7} & \textbf{81.4} & 93.1 & \underline{87.6} & \underline{85.2}  \\ 

\rowcolor{fullgreen!30} \bf \ours-CoT (Ours) &  & \textit{Lora} &  \textbf{79.3} (\textcolor{darkgreen}{\textbf{+3.7\%}}) & 81.0 (\textcolor{darkgreen}{\textbf{+6.9\%}}) & \textbf{97.1} (\textcolor{darkgreen}{\textbf{+1.4\%}}) & \textbf{88.6} (\textcolor{darkgreen}{\textbf{+2.6\%}}) & \textbf{86.5} (\textcolor{darkgreen}{\textbf{+3.6\%}}) \\ 
\bottomrule
\end{tabular}
}
\caption{\textbf{Performance comparison on FGVC datasets.} Compared to SFT (with/without CoT data) and Visual-RFT baselines, our \ours achieves the best results with consistent gains across datasets (values in parentheses denote relative improvements over the Visual-RFT (\textit{Lora})~\cite{liu2025visual} baseline). [AO: \ao prompt; SFT-CoT: SFT with CoT data; \textbf{Best} and \underline{second best} are highlighted.]}
\label{tab:main_results}
\vspace{-.5em}
\end{table*}

\section{Experiments} \label{sec:experiments}

\paragraph{Implementation Details.}
Qwen2-VL-2B-Instruct~\citep{wang2024qwen2} is used as the base model. We train with 4 NVIDIA H100 GPUs with 81G of memory. We use Qwen2-VL-7B-Instruct~\citep{wang2024qwen2} as the reward model for MLLM-based accuracy reward, and E5~\cite{wang2022text} as the embedding model for answer embedding similarity reward. To constrain the reasoning length, we set $L_{\min}{=}0$, $L_{\max}{=}10$ for \ours. We use $\gamma{=}64$ and $\alpha{=}128$ for LoRA, and a learning rate of $2e-5$ with 64 as the accumulated batch size. We set the number of generations $G{=}8$ and $\beta{=}0.04$ for GRPO. Each experiment trains with 200 step maximum, and all seeds are fixed across the training and evaluation procedures to ensure reproducibility and fairness. More details are in the supplementary.

\begin{algorithm}[t!]
\caption{Advantage Normalization with \alg}
\label{alg:advantage_norm}
\begin{algorithmic}
    \State \textbf{Input:} Number of generations $G$, number of reward functions $M$, reward matrix $\boldsymbol{R} \in \mathbb{R}^{G \times M}$, $\epsilon$.
    \State \textbf{Output:} Aggregated advantage $\boldsymbol{A} \in \mathbb{R}^{G}$.
    \Statex
    
    \Statex \textit{// 1. Compute mean and std for each reward function}
    \State $\boldsymbol{\mu} \gets \text{Mean}(\boldsymbol{R}, \text{axis}{=}0)$ \Comment{$\mathbb{R}^M$}
    \State $\boldsymbol{\sigma} \gets \text{Std}(\boldsymbol{R}, \text{axis}{=}0)$ \Comment{$\mathbb{R}^M$}
    \Statex
    
    \Statex \textit{// 2. Normalize rewards per function (element-wise)}
    \State $\boldsymbol{A}_{\text{norm}} \gets (\boldsymbol{R} - \boldsymbol{\mu}) / (\boldsymbol{\sigma} + \epsilon)$ \Comment{$\mathbb{R}^{G \times M}$}
    \Statex
    
    \Statex \textit{// 3. Aggregate scores into a final advantage}
    \State $\boldsymbol{A} \gets \text{Sum}(\boldsymbol{A}_{\text{norm}}, \text{axis}{=}1)$ \Comment{$\mathbb{R}^G$}
    \Statex
    
    \Statex \textit{// --- Original GRPO Method (for comparison) ---}
    \Statex \textit{// 1. Aggregates rewards}
    \Statex \textit{// $\boldsymbol{R}_{\text{agg}} \gets \text{Sum}(\boldsymbol{R}, \text{axis}{=}1)$}
    \Statex \textit{// 2. Normalizes the aggregated rewards}
    \Statex \textit{// $\mu_{\text{agg}} \gets \text{Mean}(\boldsymbol{R}_{\text{agg}})$; $\sigma_{\text{agg}} \gets \text{Std}(\boldsymbol{R}_{\text{agg}})$}
    \Statex \textit{// $\boldsymbol{A}_{\text{GRPO}} \gets (\boldsymbol{R}_{\text{agg}} - \mu_{\text{agg}}) / (\sigma_{\text{agg}} + \epsilon)$}    
    \State \textbf{return} $\boldsymbol{A}$
\end{algorithmic}
\end{algorithm}

\subsection{Results of \ours}

Tab.~\ref{tab:main_results} summarizes the performance across four FGVC datasets, from which several consistent patterns emerge:

\paragraph{Lora Outperforms Fully-FT.} We find that LoRA fine-tuning~\cite{hu2022lora} consistently surpasses fully fine-tuning (Fully-FT) under both SFT and RFT settings. This confirms that LoRA not only reduces computational cost but also enables the model to better leverage limited fine-grained visual data.

\paragraph{RFT Provides Stronger Gains.} Transitioning from SFT to RFT yields substantial accuracy improvements across all FGVC benchmarks. RFT enables the model to directly optimize accuracy-centric objectives, correcting undesirable generation behaviors and yielding more stable, discriminative predictions. This is especially desirable in few-shot FGVC, where precise category boundaries must be learned from limited supervision.

\paragraph{Superiority of \ours.} Building upon our findings, our proposed \ours further improves performance across all benchmarks. By integrating: (i) an ensemble, semantically-aware reward that provides dense and accuracy-aligned feedback, and (ii) the Multi-Reward Normalization module (\alg) that stabilizes heterogeneous rewards, \ours achieves consistent gains over Visual-RFT and other baselines. Importantly, \ours with only a 2B backbone and 4-shot training significantly surpasses Finedefics-8B trained on the full FGVC datasets, underscoring the efficiency and scalability of our approach.

\paragraph{Reasoning Control Outweighs Prompt Style.} 
We observe that controlling the reasoning length has a larger impact on performance than the choice of prompt style. First, No-Thinking-RFT uses an \ao-style prompt, whereas Visual-RFT uses a CoT-style prompt, yet they yield similar performance. This suggests that prompt style is not the primary performance determinant. Second, within our \ours, \ours-CoT is slightly better than the \ours-AO. Together, these observations suggest that performance depends more on reasoning-length control than on whether the prompt elicits CoT. In our view, suppressing reasoning encourages the model to focus on accuracy as the main optimization target, while still allowing it to generate reasoning when reasoning is genuinely beneficial. 
This further supports our Cost of Thinking analysis and Finding~\ref{finding1} in Sec.~\ref{sec:preliminary}, confirming that reasoning length is the key factor influencing fine-grained visual perception. We then derive the following conclusion:

\conclusion{For fine-grained visual perception, jointly using \textbf{multi-perspective, accuracy-centric rewards} and \textbf{explicit reasoning length control} leads to stronger visual perception capabilities.} \label{conclusion1}

\subsection{Ablation Studies}

\paragraph{Effects of \alg.} We employ format reward and classification reward to investigate the impact of \alg. As shown in Tab.~\ref{tab:ablation_mrpo}, integrating \alg consistently improves performance across all three FGVC datasets, yielding gains of +1.1\%/+2.1\%/+0.4\% under full fine-tuning and +0.7\%/+1.5\%/+0.6\% under LoRA. The improvements are more pronounced with larger training capacity, suggesting that \alg can better leverage additional parameters. Overall, these results confirm that \alg effectively boosts model performance while preserving strong efficacy in parameter-efficient training regimes.

\begin{figure}[t!]
    \centering
    \includegraphics[width=.95\linewidth]{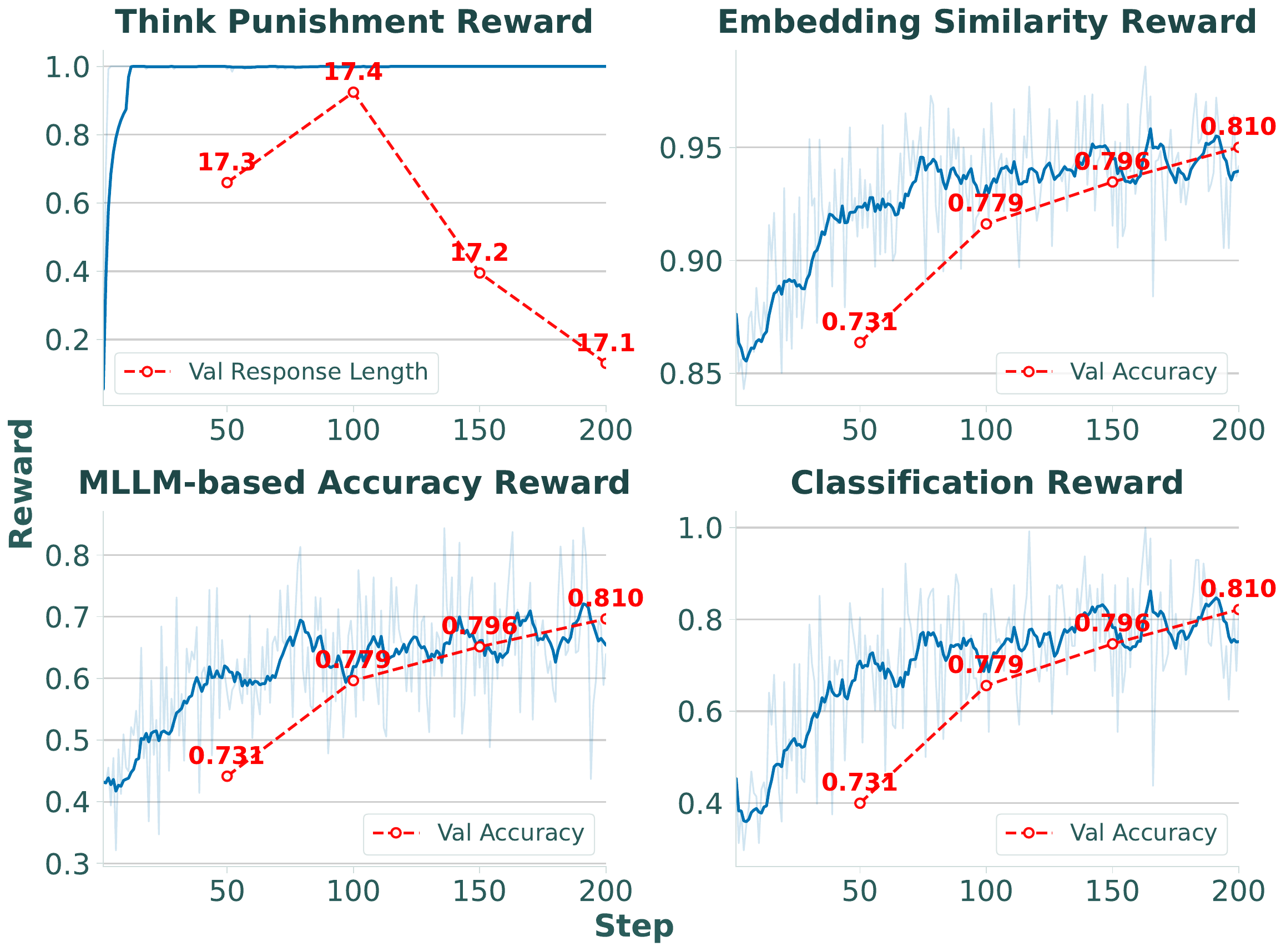}
    \caption{\textbf{Reward curves of \ours on Flowers-102.} Rewards and validation accuracy consistently increase over training, demonstrating the effectiveness of our reward design.}
    \label{fig:reward_curves_flowers_refinerft}
\vspace{-1em}
\end{figure}

\paragraph{Effects of Ensemble Reward.} We ablate the effects of the ensemble reward design in Tab.~\ref{tab:ablation_rewards}. The results demonstrate that incorporating multiple reward components leads to consistent performance gains. Each reward contributes complementary information, guiding the model toward robust learning objectives. Notably, when all reward functions are jointly combined as the ensemble reward, the model achieves the best overall performance, suggesting that aggregating complementary reward signals provides richer and more stable guidance for optimization than any individual reward. Fig.~\ref{fig:reward_curves_flowers_refinerft} shows the reward curves during training, validating the effectiveness of \ours. 

\paragraph{Effects of Trainable Parameters.} We analyze the impact of trainable parameters using the format and classification rewards on \ours. As shown in Tab.~\ref{tab:ablation_train_params}, model performance consistently improves with increasing LoRA capacity. In particular, the configuration with $r{=}64, \alpha{=}128$ achieves the highest accuracy, surpassing the Fully-FT baseline across all datasets. In contrast, the smaller setting ($r{=}16, \alpha{=}32$) leads to a notable performance drop. These results indicate that appropriately chosen LoRA rank and scaling factors can outperform fully fine-tuning in few-shot scenarios, providing an efficient and effective approach for model adaptation.

\begin{table}[t!]
\tabcolsep=0.1cm
\centering
\resizebox{0.9\linewidth}{!}{
\begin{tabular}{l|ccc}
\toprule
{\textbf{Methods}} &{\textbf{Aircrafts-102}}  &{\textbf{Flowers-102}} &{\textbf{Cars-196}} \\
\midrule
\rowcolor{gray!30} \multicolumn{4}{c}{\textit{Fully fine-tuing}} \\
GRPO~\cite{guo2025deepseek}         & 74.0 & 68.6 & 94.7 \\
\rowcolor{fullgreen!30} \bf + \alg (Ours)                   & 75.1 & 70.7 & 95.1 \\
\rowcolor{gray!30} \multicolumn{4}{c}{\textit{Lora (r=64, $\alpha$=128)}} \\
GRPO~\cite{guo2025deepseek}         & 75.6 & 74.1 & 95.7 \\
\rowcolor{fullgreen!30} \bf + \alg (Ours)       & 76.3 & 75.6 & 96.3 \\
\bottomrule
\end{tabular}
}
\caption{\textbf{Comparison of \alg under two training regimes.} \alg serves as a plug-and-play module and consistently enhances accuracy across datasets under fully fine-tuning and LoRA.}
\label{tab:ablation_mrpo}
\vspace{-.5em}
\end{table}

\begin{table}[t!]
\tabcolsep=0.1cm
\centering
\resizebox{0.9\linewidth}{!}{
\begin{tabular}{l|ccc}
\toprule
{\textbf{Methods}} &{\textbf{Aircrafts-102}}  &{\textbf{Flowers-102}} &{\textbf{Cars-196}} \\
\midrule
\textit{Fully-FT}                        & 75.1 & 70.7 & 95.1 \\
\rowcolor{gray!30} \multicolumn{4}{c}{\textit{Lora}} \\
$r=16, \alpha=32$        & 71.3 & 64.6 & 94.0 \\
$r=32, \alpha=64$        & 75.4 & 70.4 & 95.0 \\
\rowcolor{fullgreen!30} $r=64, \alpha=128$       & 76.3 & 75.6 & 96.3 \\
\bottomrule
\end{tabular}
}
\caption{\textbf{Comparison of trainable parameters.} LoRA with larger ranks ($r$) and scaling factors ($\alpha$) progressively improves accuracy, eventually surpassing full fine-tuning across all datasets.}
\label{tab:ablation_train_params}
\vspace{-.5em}
\end{table}

\begin{table}[t!]
    \tabcolsep=0.1cm
    \centering
    \resizebox{0.88\linewidth}{!}{%
    \begin{tabular}{ccccc|cc}
    \toprule
     $R_f$ & $R_{cls}$ & $R_{len}$ & $R_{mllm}$ & $R_{emb}$ & Aircrafts-102 & Pets-37 \\
     \midrule
        \checkmark & \checkmark & & & & 76.3 & 86.8 \\
        \checkmark & \checkmark & \checkmark & & & 78.5 &  87.5 \\
        \checkmark & \checkmark & \checkmark & \checkmark & & 77.5 & 85.7 \\
        \checkmark & \checkmark & \checkmark & & \checkmark& 79.0 & 86.3 \\
        \rowcolor{fullgreen!30} \checkmark & \checkmark & \checkmark & \checkmark & \checkmark & 79.3 & 88.6\\
    \bottomrule
    \end{tabular}
    }%
    \caption{\textbf{Effects of ensemble reward.} Combining multiple reward functions consistently improves performance, and using all rewards yields the best overall results.}
    \label{tab:ablation_rewards}
\vspace{-.5em}
\end{table}

\begin{figure}[t!]
    \centering
    \includegraphics[width=.9\linewidth]{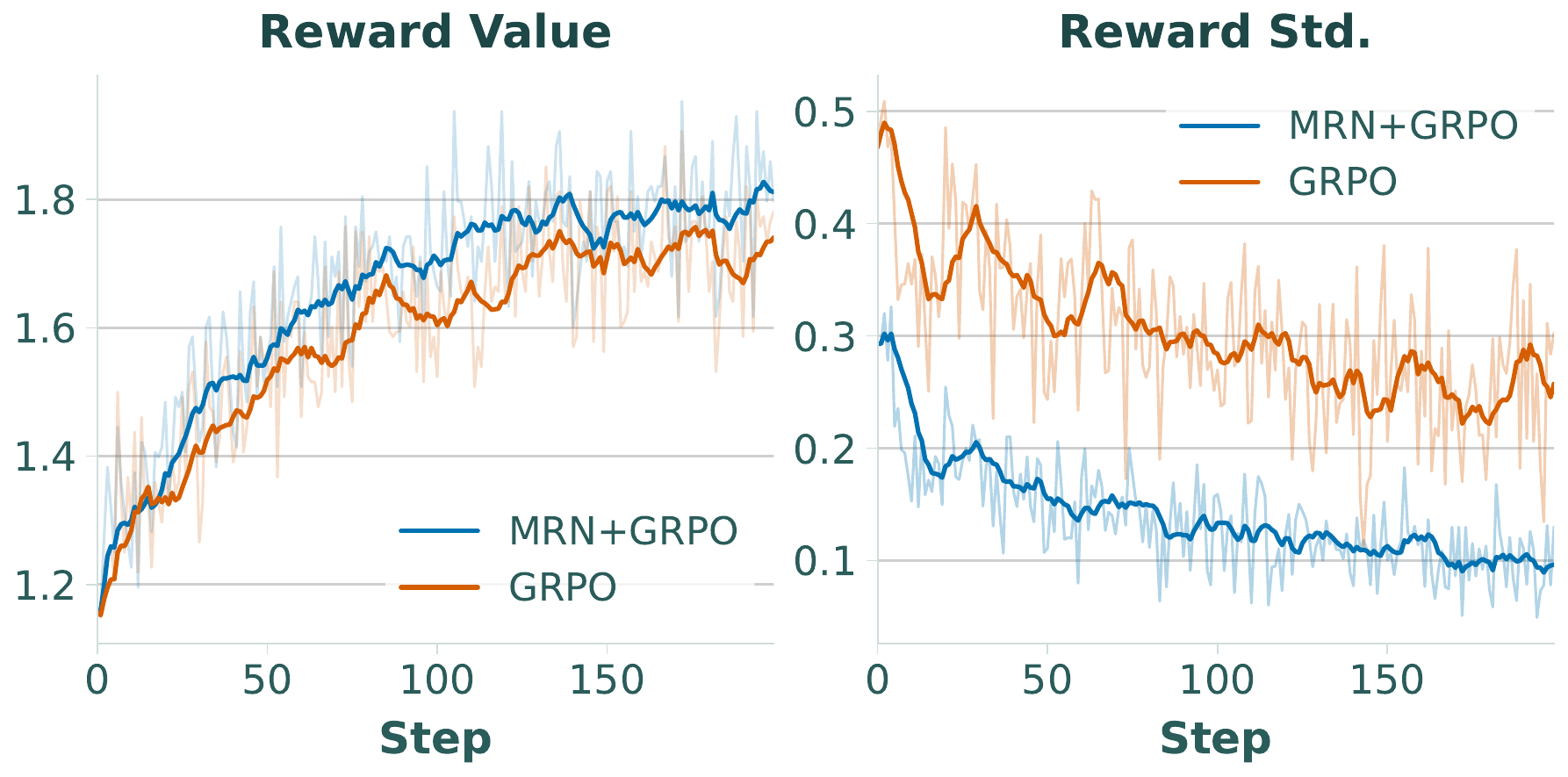}
    \caption{\textbf{Training reward and its standard deviation comparison on Aircrafts-102.} $\text{\alg}+\text{GRPO}$ achieves consistently higher reward values and lower variance throughout training, indicating improved stability and optimization efficiency.} 
    \label{fig:reward_comparison_aircrafts}
\vspace{-.5em}
\end{figure}

\paragraph{Reward Distribution Comparison.} As shown in Fig.~\ref{fig:reward_comparison_aircrafts}, our proposed \alg consistently achieves higher reward values and maintains significantly lower reward standard deviation compared to the baseline GRPO. Throughout training, \alg exhibits a steady improvement in reward, indicating more stable and efficient policy optimization. In contrast, GRPO shows larger standard deviation, reflecting less stable learning behavior. The notably lower reward variance of \alg suggests that it effectively mitigates gradient noise and reduces policy fluctuation, leading to smoother and more reliable reward progression. These observations demonstrate that \alg not only enhances training stability but also enables more consistent reward maximization, thereby improving optimization robustness and efficiency.

\begin{figure}[t!]
    \centering
    \includegraphics[width=\linewidth]{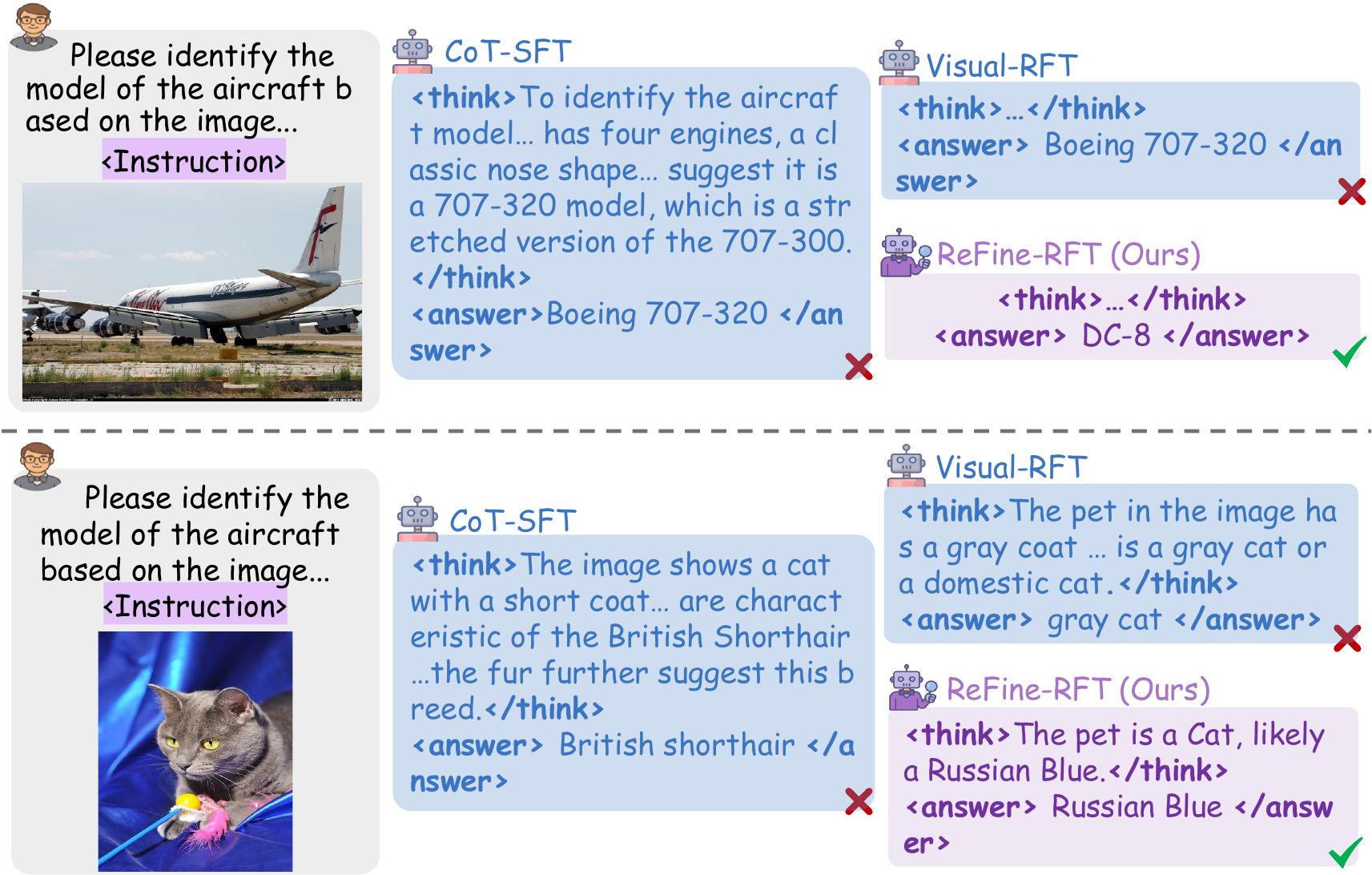}
    \caption{\textbf{Comparison of responses.} SFT-CoT and Visual-RFT produce long reasoning with incorrect answers, while \ours achieves concise reasoning and higher accuracy. More results and analyses are in the supplementary.}  
    \label{fig:cases_comparison}
\vspace{-1em}
\end{figure}

\paragraph{Qualitative Results.} As shown in Fig.~\ref{fig:cases_comparison}, \ours demonstrates a clear advantage in both reasoning efficiency and accuracy, encouraging minimal reasoning steps.

\section{Conclusion} \label{sec:conclusion}

We reveal the ``Cost of Thinking'' in FGVC for MLLMs, showing that excessive textual reasoning degrades fine-grained visual perception performance from both the inference and training perspectives. Our systematic study across zero-shot and multiple fine-tuning regimes indicates that perception-centric tasks benefit more from concise reasoning. Motivated by this, we propose \ours, a reasoning-constrained RFT framework that enhances visual perception by integrating ensemble, semantically-aware rewards with a Multi-Reward Normalization (\alg) for optimization. Extensive results demonstrate that \ours achieves state-of-the-art performance across FGVC benchmarks, highlighting that effective visual perception emerges from constraint thinking and accuracy-centric reward shaping. Future work will probe the mechanisms behind the Cost of Thinking and extend \ours to broader multimodal tasks.



{\small
\bibliographystyle{ieee_fullname}
\bibliography{main}
}

\clearpage
\clearpage
\setcounter{page}{1}
\maketitlesupplementary

\section{GRPO Algorithm}

GRPO requires the model to sample $G$ diverse responses $\{o_1, o_2, \dots, o_G\}$ from the current model $\pi_{\theta}$ and obtains rewards $\{r_1, r_2, \dots, r_G\}$ for $o_i$. GRPO assesses the relative quality by normalizing $r_i$ using the mean and standard deviation of the group reward (details provided in the main paper). With the group normalization, GRPO encourages the model to sample preferred answers with a higher reward. The model is updated via:
\begin{equation}
\begin{aligned}
    J_{\mathrm{GRPO}}(\theta) 
    &= \mathbb{E}_{q \sim P(Q),\, \{o_i\}_{i=1}^G \sim \pi_{\theta_{\mathrm{old}}}(O \mid q)} \\
    &\qquad\quad  \Bigg[ 
        \frac{1}{G} \sum_{i=1}^G 
        \min \Bigg( 
            \frac{\pi_{\theta}(o_i \mid q)}{\pi_{\theta_{\mathrm{old}}}(o_i \mid q)} A_i, \\
    &\qquad\quad 
            \mathrm{clip}\!\left(
                \frac{\pi_{\theta}(o_i \mid q)}{\pi_{\theta_{\mathrm{old}}}(o_i \mid q)}, 
                1 - \varepsilon, 
                1 + \varepsilon
            \right) A_i
        \Bigg) \\ 
    &\qquad\quad 
    - \beta \, D_{\mathrm{KL}}\!\big( \pi_{\theta} \parallel \pi_{\mathrm{ref}} \big)
    \Bigg]
\end{aligned}
\end{equation}
where $\varepsilon$ and $\beta$ are the GRPO clipping hyperparameters and the coefficient weight for controlling the Kullback–Leibler (KL) penalty~\citep{schulman2017proximal}, respectively. $\pi_\mathrm{ref}$ is the reference model.

\section{Additional Implementation Details}

\subsection{Datasets}
\paragraph{Statistics of Training and Evaluation Set.}
We use the 4-shot data provided by~\cite{liu2025visual}. The statistics of the training set and evaluation set can be found in Tab.~\ref{tab:dataset_stat_supp}.

\begin{table}[t!]
  \centering
  \resizebox{\linewidth}{!}{%
  \begin{tabular}{lcccc}
    \toprule
    Dataset        & \#Categories & Train & 4-shot (\%) & Test \\ 
    \midrule
    Aircraft\textendash102  & 100 & 3\,334 & 400 (12.0\%) & 3\,333 \\ 
    Flower\textendash102    & 102 & 1\,020 & 408 (40.0\%) & 2\,463 \\ 
    Pet\textendash37        & 37  & 3\,680 & 148 (4.0\%) & 3\,669 \\ 
    Car\textendash196       & 196 & 8\,144 & 784 (9.6\%) & 8\,041 \\ 
    \bottomrule
  \end{tabular}
  }%
  \caption{Statistics of FGVC datasets. The ``4-shot'' column shows the number of images we used for training.}
  \label{tab:dataset_stat_supp}
\end{table}

\paragraph{Prompts.} Fig.~\ref{fig:choice_prompt} and Fig.~\ref{fig:cot_prompt} show the prompts for \ao~and \cott, respectively, while Fig.~\ref{fig:judge_prompt} provides the prompt for the MLLM-based accuracy reward. The \ao~prompt is used for SFT training, and the \cott~prompt for both CoT-SFT and RFT training. Placeholders {DATASET}, {PRED}, and {GT} are used to denote the specific dataset name (\eg, plants, aircrafts), the model's predicted label, and the ground truth label, respectively.

\begin{figure}[t]
    \begin{tcolorbox}[parbox=false,colback=fullgreen!10, colframe=fullgreen!50, title= \ao Prompt, coltitle=black]
    This is an image containing an \{DATASET\}. Please identify the \{DATASET\} of the \{DATASET\} based on the image. Only provide the final answer directly, without any explanation or special formatting.
    \end{tcolorbox}    
    \caption{\ao prompt.}
    \label{fig:choice_prompt}
\end{figure}

\begin{figure*}[t]
    \begin{tcolorbox}[parbox=false,colback=fullgreen!10, colframe=fullgreen!50, title=\cott Prompt, coltitle=black]
    This is an image containing an \{DATASET\}. Please identify the \{DATASET\} of the \{DATASET\} based on the image. Output the thinking process in \think and final answer in \answer tags. The output answer format should be as follows: \texttt{<think>...</think> <answer>species name</answer>}. Please strictly follow the format."    
    \end{tcolorbox}
    \caption{\cott prompt.}
    \label{fig:cot_prompt}
\end{figure*}

\begin{figure*}[t]
    \begin{tcolorbox}[size=small, parbox=false,colback=fullgreen!10, colframe=fullgreen!50, title=Judge Prompt, coltitle=black]
    You are a scoring assistant. Based on the similarity between the "Predicted Answer" and the "Correct Answer", provide a score from 0 to 10. A score of 10 means a perfect match, and 0 means a complete mismatch. You must output only the numerical score.

    ---
    
    [Example 1]
    
    Predicted Answer: "2007 Dodge Dakota Club Cab"
    
    Correct Answer: "2007 Dodge Dakota Club Cab"
    
    Score: 10
    
    ---
    
    [Example 2]
    
    Predicted Answer: "Boeing 707"
    
    Correct Answer: "707-320"
    
    Score: 6
    
    ---
    
    [Example 3]
    
    Predicted Answer: "Nasturtium"
    
    Correct Answer: "watercress"
    
    Score: 0
    
    ---
    
    [Your Task]
    
    Predicted Answer: "\{PRED\}"
    
    Correct Answer: "\{GT\}"
    
    Score:
    
    \end{tcolorbox}
    \caption{Judge prompt for MLLM-based accuracy reward.}
    \label{fig:judge_prompt}
\end{figure*}

\paragraph{Reward Model Implementation.} We deploy Qwen2-VL-7B~\cite{wang2024qwen2} as the reward model for MLLM-based accuracy reward using LMDeploy~\cite{2023lmdeploy}. To optimize performance, we employ mixed precision and the TurboMind inference backend. LMDeploy provides a flexible framework with OpenAI-compatible APIs, ensuring broad compatibility and facilitating the potential integration of other teacher models in the future.

\begin{figure}[t!]
    \centering
    \includegraphics[width=\linewidth]{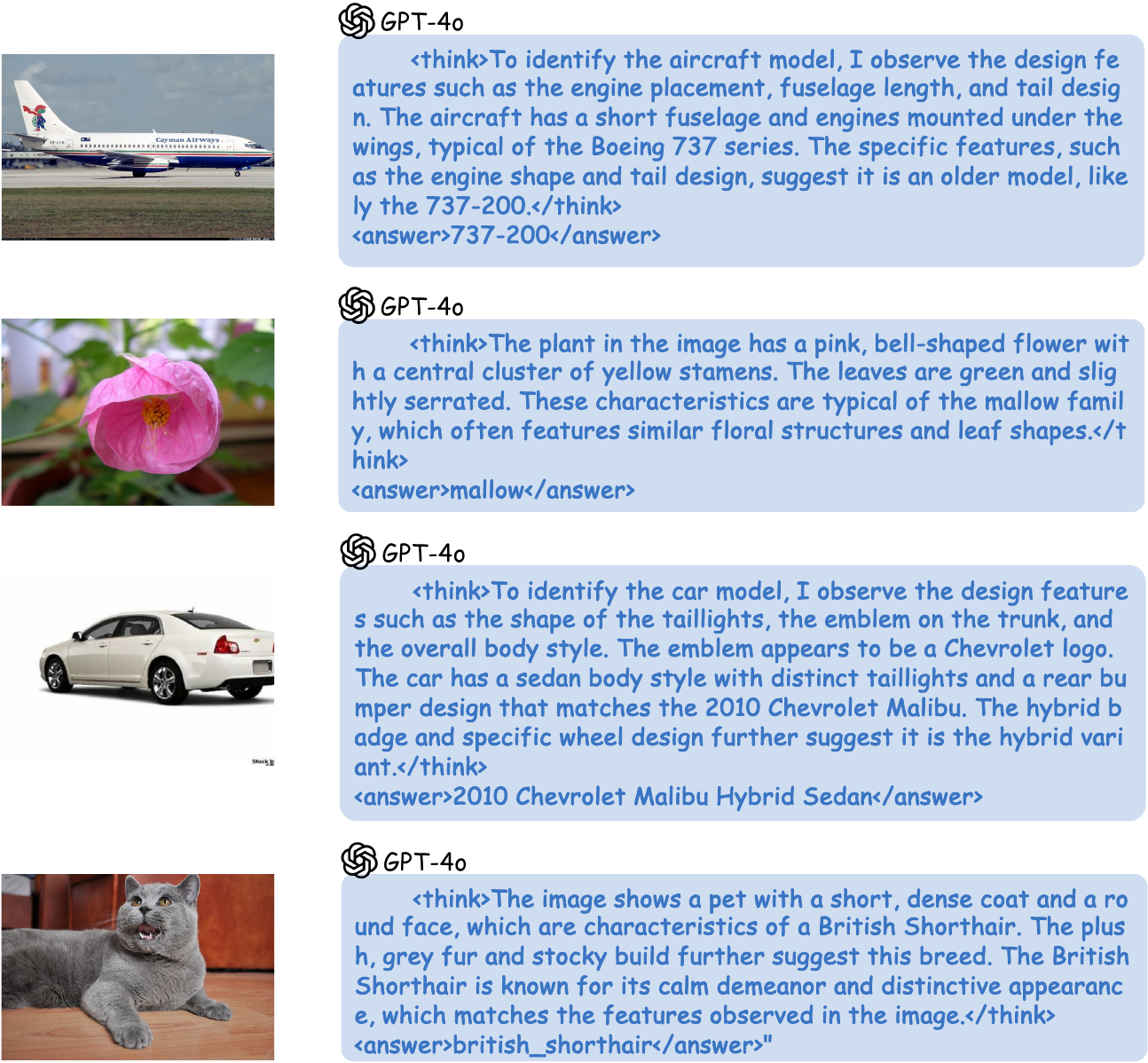}
    \caption{Examples of CoT annotations generated by GPT-4o, featuring long and fine-grained reasoning traces that describe each image in detail.}
    \label{fig:cot_dataset_example}
\end{figure}

\paragraph{CoT Data Curation.} We employ GPT-4o-2024-08-06~\cite{hurst2024gpt} to generate high-quality Chain-of-Thought data. For each sample, we provide the image, question prompt, and corresponding ground truth label, instructing the model to generate reasoning that leads to the correct answer. This ensures the accuracy of the synthesized CoT data. The prompt template shown in Fig.~\ref{fig:sft_cot_prompt} uses SOLUTION as a placeholder for the ground truth label, while Fig.~\ref{fig:cot_dataset_example} displays representative examples of the generated data.

\begin{figure*}[t]
    \begin{tcolorbox}[size=small, parbox=false,colback=fullgreen!10, colframe=fullgreen!50, title=SFT-CoT Prompt, coltitle=black]
    This is an image containing a pet. Please identify the species of the pet based on the image.
    Output the thinking process in \texttt{<think> </think>} and final answer in \texttt{<answer> </answer>} tags.The output answer format should be as follows:
    \texttt{<think> ... </think> <answer>species name</answer>}
    Please strictly follow the format. The ground truth answer is \{SOLUTION\}. Limit your response to 100 words.
    \end{tcolorbox}
    \caption{Judge prompt for MLLM-based accuracy reward.}
    \label{fig:sft_cot_prompt}
\end{figure*}

\paragraph{Additional Training Implementation Details.} To ensure reproducibility, all experiments use fixed random seeds. We employ BF16 precision and apply LoRA with a rank $\gamma=64$ and scaling parameter $\alpha=128$ to the following modules: \verb|q_proj|, \verb|k_proj|, \verb|v_proj|, \verb|o_proj|, \verb|gate_proj|, \verb|up_proj|, \verb|down_proj|. Models are trained for a maximum of 200 steps with a completion length capped at 256 tokens.

\paragraph{Performance Validation Details.} Performance is evaluated exclusively by answer accuracy. For \cott, we follow~\cite{liu2025visual} to extract answers from the \answer\ tag. A prediction is considered correct if a normalized substring match exists in either direction between the extracted answer and the ground truth. For \ao, responses are evaluated directly, as their output format is inherently comparable to the ground truth.

\section{Additional Experimental Results}

\paragraph{Additional results on prompt-type choices for RFT.}
We further study how the prompt type used during RFT affects performance. Following Visual-RFT~\cite{liu2025visual}, we train both the chain-of-thought (\cott) and answer-only (\ao) variants under the same setting. As reported in Tab.~\ref{tab:supp_ablation_prompt_diff}, the two variants achieve almost identical accuracies on Aircrafts-102 and Cars-196, suggesting that explicitly generating long CoT traces brings little additional benefit beyond an answer-only prompt, which is consistent with our comparison between Visual-RFT~\cite{liu2025visual} and No-Thinking-RFT~\cite{li2025think}.

\begin{figure}
    \centering
    \includegraphics[width=0.9\linewidth]{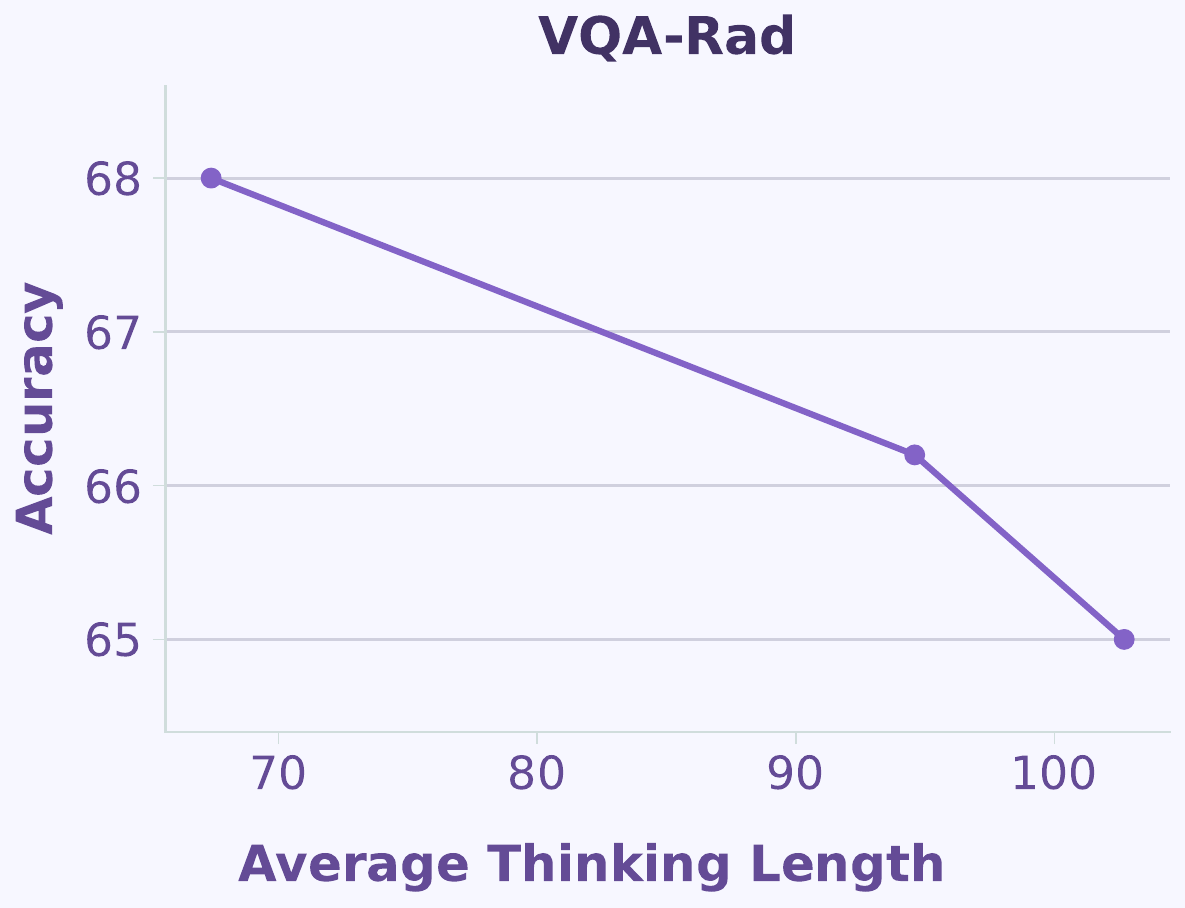}
    \caption{Cost of thinking on VQA-Rad dataset. Performance decreases as the thinking length increases.}
    \label{fig:cot_vqa-rad}
\end{figure}

\paragraph{Extending to Other FGVC Tasks.} 
Fig.~\ref{fig:cot_vqa-rad} shows that the Cost of Thinking exists in the medical imaging dataset VQA-Rad.

\paragraph{Comparisons against discriminative models.}
We show the comparison in Tab.~\ref{tab:rebuttal_clip_performance}. Compared with CLIP and its discriminative variants, our method achieves the best performance on Cars and Aircrafts, improving over CLIP$^{\text{LP}}$ by a large margin (+10.4\% and +19.8\%, respectively). This suggests that our method is particularly effective on fine-grained categories with subtle inter-class differences. On Flowers and Pets, however, linear probing on CLIP remains more competitive, indicating that discriminative adaptation is still advantageous on domains with relatively cleaner visual cues or stronger alignment to CLIP pre-training. Overall, these results show the promise of MLLMs on fine-grain visual understanding, and our method is highly competitive with standard discriminative baselines.

\begin{table}[t!]
\tabcolsep=0.1cm
\centering
\resizebox{\linewidth}{!}{
\begin{tabular}{c|cccc}
Methods & Cars & Aircrafts & Flowers & Pets\\
\midrule
CLIP (ViT-B/16)$^*$                 & 65.6 & 27.1 & 70.4 & 88.9 \\
$\text{CLIP}^{\text{LP}}$ (ViT-B/16)$^*$       & 86.7 & 59.5 & 98.1 & 93.1 \\
$\text{CLIP}^{\text{SimNL}}$ (4-shot)   & 68.0 & 29.0 & 92.0 & 88.1 \\
Ours                    & 97.1 & 79.3 & 81.0 & 88.6 \\
\end{tabular}
}
\vspace{-.5em}
\caption{Comparison with discriminative models. $^*$: from CLIP official report. LP: Linear Probing. SimNL:~\cite{zhang2025enhancing}.}
\label{tab:rebuttal_clip_performance}
\vspace{-1em}
\end{table}

\begin{figure}
    \centering
    \includegraphics[width=0.9\linewidth]{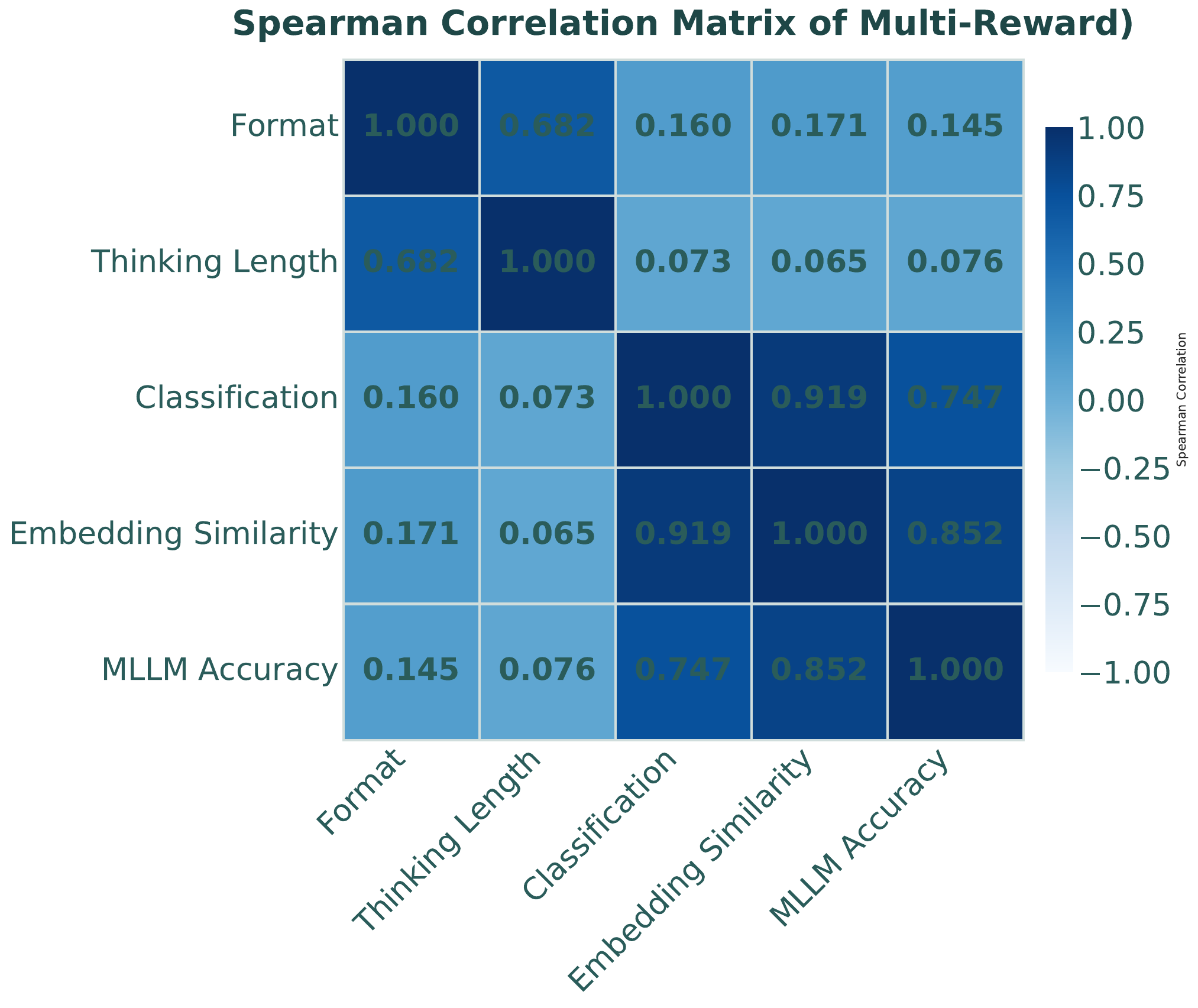}
    \caption{Correlation heatmap of ensemble rewards.}
    \label{fig:spearman_heatmap}
\end{figure}

4\paragraph{Correlation Analysis of Rewards.}
As shown in Tab.~\ref{tab:ablation_rewards} and Fig.~\ref{fig:reward_curves_flowers_refinerft}, combining all rewards yields the best performance, and $R_{\text{cls}}$, $R_{\text{emb}}$, and $R_{\text{mllm}}$ show consistent positive trends. Fig.~\ref{fig:spearman_heatmap} on the Flowers test set further shows that the rewards are correlated yet distinct. This indicates that these three rewards are aligned in encouraging semantically correct predictions, but they are not redundant and still provide complementary learning signals. By contrast, the format reward and thinking-length reward have much weaker correlations with the task-related rewards, suggesting that they mainly regularize output structure and response behavior rather than directly optimizing classification performance. Interestingly, the format reward is relatively correlated with thinking length, implying that shorter reasoning often co-occurs with better-formatted responses. 

\begin{figure}[t!]
    \centering
    \includegraphics[width=\linewidth]{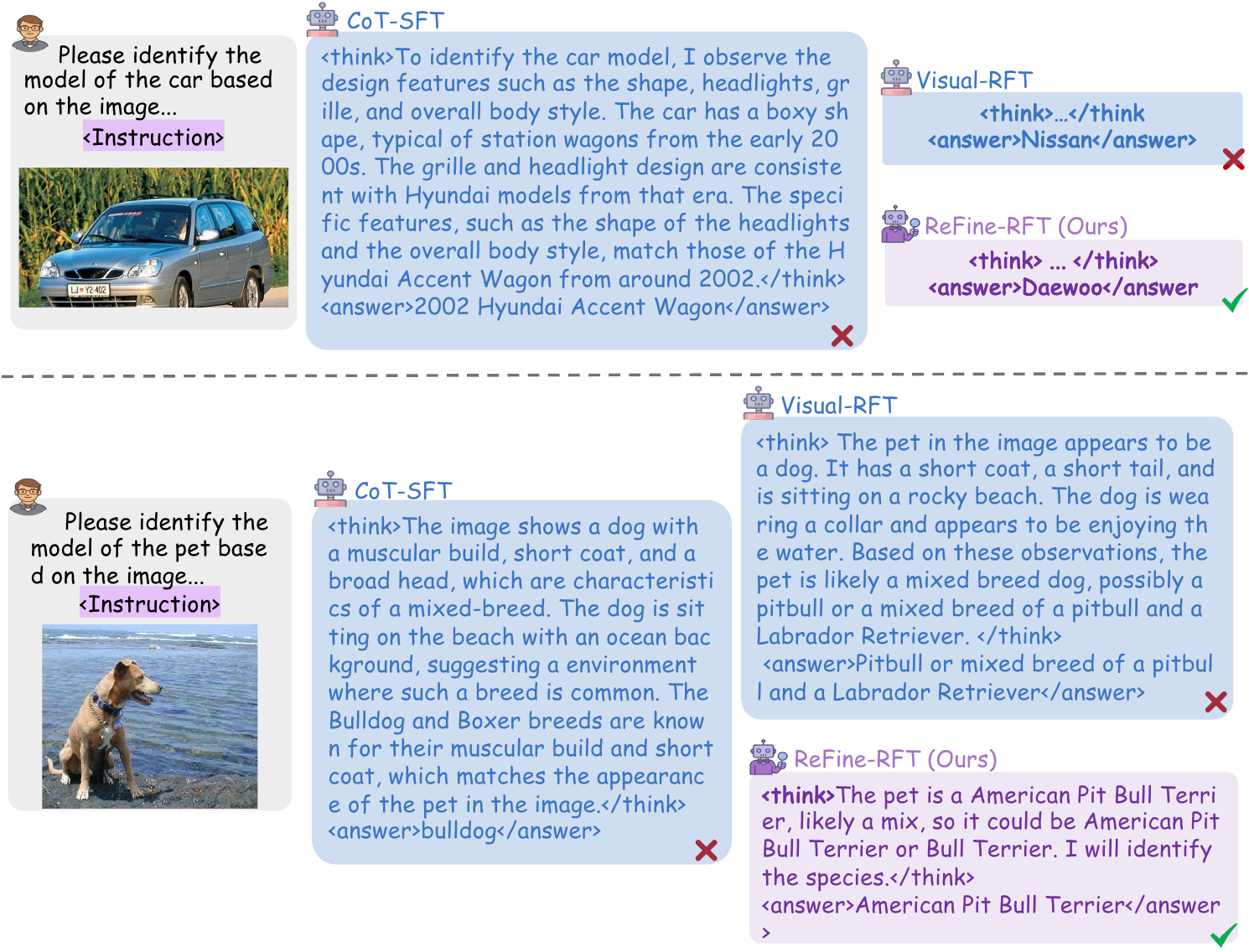}
    \caption{\textbf{Additional comparison of responses.}}
    \label{fig:supp_cases_comparison}
\end{figure}

\paragraph{Additional Qualitative Results.} We provide additional qualitative results in Fig.~\ref{fig:supp_cases_comparison}.

\paragraph{Thinking Length comparison.} Across all four datasets, there is a clear and consistent ordering of thinking lengths: SFT-CoT produces the longest chains of thought, as SFT-CoT data contains long reasoning traces. Zero-shot sits in the middle, while Visual-RFT substantially shortens the reasoning, and \ours is the most concise. The gap is especially striking on Cars and Aircrafts, where SFT-CoT more than doubles or even triples the thinking length of \ours. Combined with our empirical observation that training with longer thinking lengths actually hurts task performance, this pattern suggests that excessive CoT introduces redundancy and noise rather than useful intermediate supervision. Long SFT-CoT traces likely contain distracted or unhelpful information, which dilutes the gradient signal and encourages the model to mimic verbosity instead of learning the decision-critical answering. Zero-shot, which is not explicitly trained to be verbose, yields somewhat shorter traces and better aligns with test-time behavior, but still carries uncontrolled overthinking. This is possibly because of the pretraining data distribution. In contrast, \ours explicitly regularize the model toward concise, high-utility rationales: we focus on accuracy-centric signals that are tied to the final prediction and constrain reasoning tokens. This not only reduces token cost, but empirically correlates with higher accuracy, suggesting that there is an optimal, concise reasoning horizon, and that pushing the model to produce ever-longer CoT drives it into a worse visual perception performance.

\begin{figure}
    \centering
    \includegraphics[width=\linewidth]{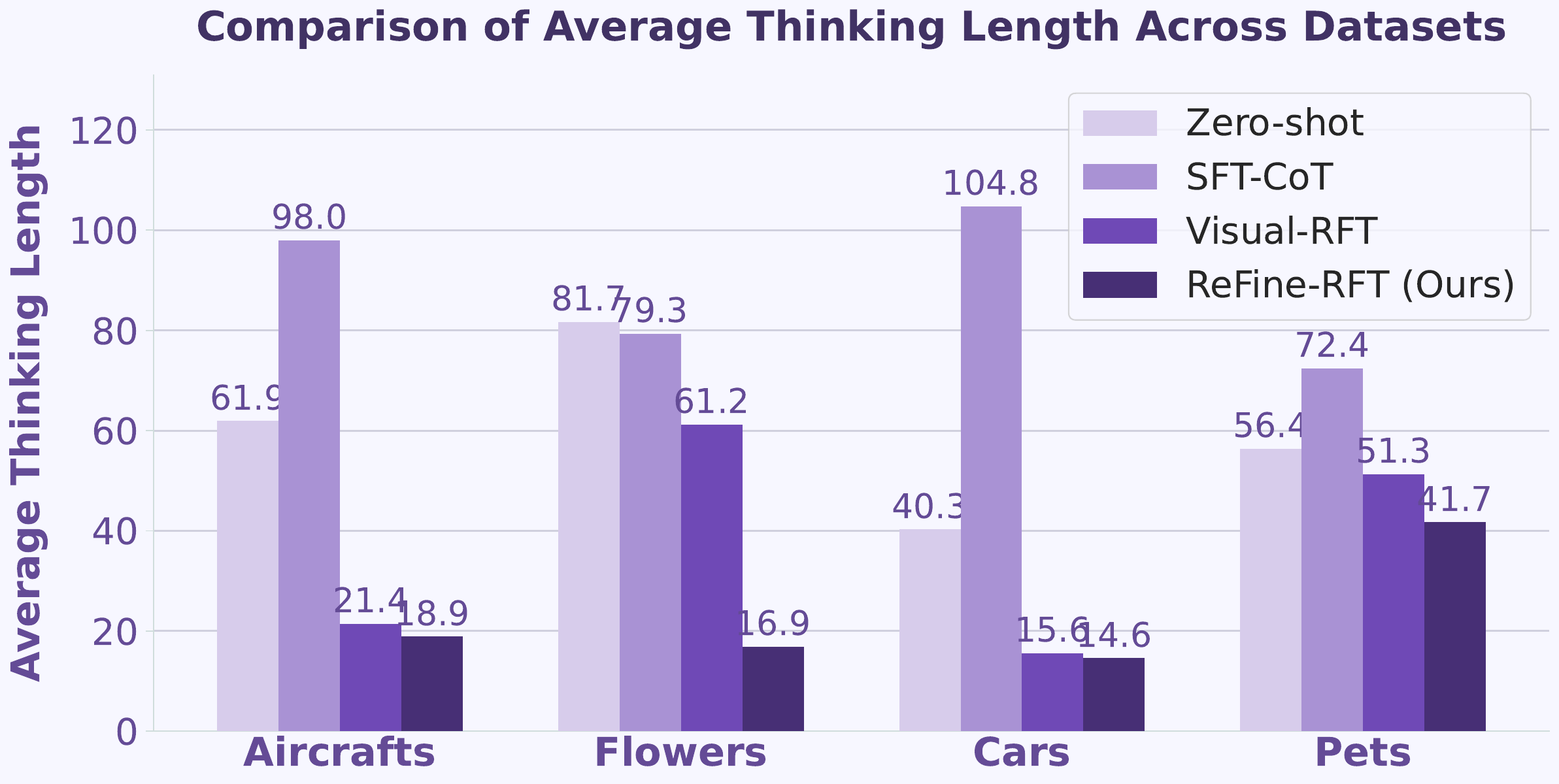}
    \caption{\textbf{Comparison of average thinking length across datasets.}
    We average the number of thinking tokens as the thinking length per dataset. SFT-CoT consistently yields the longest chains of thought, Zero-shot produces medium-length traces, while both Visual-RFT and especially ReFine-RFT generate much more concise reasoning. Our method attains the shortest thinking length on all datasets, indicating that strong performance does not require long reasoning traces.}
    \label{fig:thinking_length_comparison}
\end{figure}

\begin{table}[t!]
\tabcolsep=0.1cm
\centering
\resizebox{0.9\linewidth}{!}{
\begin{tabular}{l|cc}
\toprule
{\textbf{Methods}} &{\textbf{Aircrafts-102}}  &{\textbf{Cars-196}} \\
\midrule
\bl-AO      & 75.8 & 95.8 \\
\bl-CoT     & 75.6 &  95.7 \\

\bottomrule
\end{tabular}
}
\caption{\textbf{Comparison of prompt types in Visual-RFT.} Using an \ao prompt (Visual-RFT-AO) attains almost identical accuracy to using an explicit \cott prompt (Visual-RFT-CoT), indicating that long reasoning traces are not necessary for effective RFT on these benchmarks.}

\label{tab:supp_ablation_prompt_diff}
\vspace{-.5em}
\end{table}

\section{Potential Reasons of CoT Degradation on Visual Tasks.}

We hypothesize that the observed ``Cost of Thinking'' arises from two interacting effects. First, long textual chains-of-thought may compete with visual processing for the model’s finite attention and context budget: as more self-generated tokens accumulate, the transformer increasingly attends to its own linguistic history rather than the image embeddings, amplifying language priors while suppressing subtle visual cues that are crucial for FGVC. A closely related “attention diversion’’ phenomenon has been documented in instruction-following LLMs, where explicit CoT reduces focus on constraint tokens and significantly harms compliance accuracy~\cite{li2025thinking}, and in multimodal reasoning, where reasoning primarily in the language space leads to strong language bias and under-utilization of image features, motivating architectures that explicitly replay or re-ground visual information during reasoning~\cite{wang2025vgr, prabhakar2024deciphering}.

Second, extending the CoT sequence increases exposure to noisy or unfaithful reasoning: each additional step is an opportunity to introduce hallucinated details, spurious correlations, or incorrect intermediate visual descriptions, which are then propagated and rationalized downstream. Prior analyses of CoT on text-only tasks have shown that explanations are often unfaithful to the model’s true decision process and can rationalize biased or incorrect predictions~\cite{lanham2023measuring, ye2022unreliability}, and that error rates grow with the number of implicit reasoning operations, consistent with a “noisy reasoning’’ view where longer chains accumulate more mistakes. In fine-grained visual classification, where decisions hinge on subtle, localized perceptual evidence, such mis-grounded or noisy chains are particularly detrimental: once the CoT commits to an incorrect local description (e.g., misidentifying a part or texture), subsequent reasoning tends to reinforce that error instead of revisiting the image, making verbose CoT systematically worse than concise, answer-focused predictions.

\section{Potential Social Impact}

Our work advances the reasoning and fine-grained recognition capabilities of MLLMs, with the potential to significantly benefit real-world applications in domains such as biodiversity monitoring, medical diagnostics, industrial inspection, and scientific research, where expert-level fine-grained categorization is crucial. By enabling MLLMs to generate interpretable reasoning steps in addition to accurate predictions, our method promotes transparency and trustworthiness, critical factors for safe AI deployment in high-stakes environments. We believe this research contributes to the broader goal of making MLLMs more reliable, interpretable, and aligned with human values, while acknowledging the necessity of continuous ethical scrutiny as these systems become increasingly capable.

\section{Limitation}

While \ours\ achieves strong performance on FGVC, several limitations remain. 
First, although suppressing excessive thinking length indirectly improves training efficiency, the overall RFT pipeline is still more time-consuming than standard SFT due to the rollout sampling strategy and RFT optimization. Second, our analysis mainly focuses on the impact of thinking length and the comparison between SFT-AO, SFT-CoT, and our RFT variants, whereas the effects of \emph{thinking quality} during the RFT process remain unexplored. 
Finally, we conduct a detailed study only on fine-grained visual classification (FGVC); extending our framework and analyses to other visual tasks such as object detection, visual grounding, or more open-ended vision–language reasoning is an important direction for future work.

\end{document}